\documentclass{vki}

\usepackage{lmodern,pdftexcmds,amsmath}

\usepackage{amsfonts,amsthm,bm} % Math packages

\DeclareMathAlphabet{\mathsfbi}{OT1}{\sfdefault}{bx}{sl}
\DeclareMathVersion{sfletters}
\SetSymbolFont{letters}{sfletters}{OML}{ntxsfmi}{b}{it}

\makeatletter
\newcommand{\mathbfsbilow}[1]{%
	\text{\mathversion{sfletters}$\m@th#1$}%
}

\DeclareRobustCommand{\tensor}[1]{%
	\begingroup
	\ifcat\noexpand #1\relax
	\edef\greek@test{\detokenize{#1}}%
	\edef\greek@test{\expandafter\@cdr\greek@test\@nil}%
	\edef\greek@test{\expandafter\@car\greek@test\@nil}%
	\edef\x{\the\lccode\expandafter`\greek@test}%
	\edef\y{\number\expandafter`\greek@test}%
	\ifnum\x=\y\relax
	\mathbfsbilow{#1}%
	\else
	\mathsfbi{#1}%
	\fi
	\else
	\mathsfbi{#1}%
	\fi
	\endgroup
}
\makeatother

\usepackage[most]{tcolorbox}
\usepackage{booktabs}
\usepackage{amsmath}
\usepackage{wrapfig}
\usepackage[linesnumbered,ruled,vlined]{algorithm2e}% http://ctan.org/pkg/algorithm2e
\usepackage{algpseudocode}
\usepackage{pdflscape}
\usepackage{subcaption}

\usepackage{siunitx}
\usepackage{comment}

\newcommand{\argmin}{\operatornamewithlimits{argmin}}

\usepackage{mathtools}
\tcbset{enhanced,colback=cyan!2!white,colframe=cyan!90!white,coltitle=blue!20!black,fonttitle=\bfseries}

\usepackage{listings}
\usepackage{hyperref}
\hypersetup{
	colorlinks=true,
	linkcolor=blue,
	filecolor=magenta,      
	citecolor=blue,
	urlcolor=cyan,
}

\begin{document}

% USER ENTRY ON
%\layout
% uncomment this \layout to have an idea of the margins
% USER ENTRY OFF

\pagenumbering{roman}
% USER ENTRY ON
\title{Fundamentals of Regression (Chapter 2)}
\author{Miguel A. Mendez\thanks{mendez@vki.ac.be}\\von Karman Institute for Fluid Dynamics}

\date{3 December 2024} % I suggest you adjust this manually
% USER ENTRY OFF
\maketitle

This chapter opens with a review of classic tools for regression, a subset of machine learning that seeks to find relationships between variables. With the advent of scientific machine learning\footnote{The term “scientific machine learning” refers to the integration of machine learning with models, principles, and data arising from the natural sciences and engineering. It does not imply that other forms of machine learning are non-scientific. Instead, it highlights a focus on problems where physical laws (e.g., conservation laws, PDEs, thermodynamics) play a central role. Scientific machine learning typically involves combining data-driven methods with domain knowledge, physics-based modeling, numerical simulation, and uncertainty quantification. The emphasis is on developing algorithms that are constrained by—or informed by—scientific theory, enabling improved prediction, interpretability, and generalization in complex physical systems.} this field has moved from a purely data-driven (statistical) formalism to a constrained or ``physics-informed'' formalism, which integrates physical knowledge and methods from traditional computational engineering. In the first part, we introduce the general concepts and the statistical flavor of regression versus other forms of curve fitting. We then move to an overview of traditional methods from machine learning and their classification and ways to link these to traditional computational science. Finally, we close with a note on methods to combine machine learning and numerical methods for physics 

\vspace{30mm}

How to cite this work
\bigskip

\begin{centering}
	\begin{lstlisting}
@InCollection{Mendez2024,
  author    = {Mendez, Miguel A.},
  title     = {Fundamentals of Regression},
  booktitle = {Machine Learning for Fluid Dynamics},
  editor    = {Mendez, Miguel A. and Parente, Alessandro},
  publisher = {von Karman Institute},
  year      = {2024},
  chapter   = {2},
  isbn      = {978-2875162090}
}
	\end{lstlisting}
\end{centering}

%%%%%%%%%%%%%%%%%%%%%%
\pagenumbering{arabic}
\setcounter{page}{1}
\clearpage{\pagestyle{empty}} 

\tableofcontents
\clearpage{\pagestyle{empty}} 

\vspace{-3mm}

\section{A note on notation and style}\label{ch_2_sec_0}

\textbf{Vectors, Matrices and lists}. We use lowercase letters for scalar quantities, i.e. $a\in\mathbb{R}$. Bold lowercase letters are used for vectors, i.e., $\bm{a}\in\mathbb{R}^{n_a}$. 
The i-th entry of a vector is denoted with a subscript as $\bm{x}_i$ or with Python-like notation as $\bm{x}[i]$. We use square brackets to create vectors from a set of scalars, e.g $\bm{a}=[a_0,a_1,\dots a_{n_a-1}]\in\mathbb{R}^{n_a}$.
Unless otherwise stated, a vector is a column vector. The use of transposition when defining a vector embedded within the text of a paragraph or sentence (inline) is omitted. 
\\We use the upper case bold letters for matrices, e.g., $\bm{A}\in {\mathbb{R}^{n_r\times n_c}}$, with $n_r$ the number of rows and $n_c$ the number of columns. The matrix entry at the i-th row and j-th column are identified as $\bm{A}_{i,j}$ or with the Python-like notation as $\bm{A}[i,j]$. When the Python notation is used, the indices begin with $0$.
We occasionally work with lists of quantities. Following a Python notation, we enclose lists within round brackets and use bold letters for a list of vectors or matrices, e.g. $\bm{\Gamma}_*=(\bm{x},\bm{y})$.

\textbf{Functions and Calculus}. Lowercase letters followed by parentheses indicate functions, regardless of whether these are scalar or vector-valued functions, i.e. $a(\bm{x}): \mathbb{R}^{n_i}\rightarrow \mathbb{R}^{n_o}$.  In a vector value function, $\bm{y}=f(\bm{x})$ and the subscript is used to define the mapping to each component, i.e. $\bm{y}_i= f_i (\bm{x})$. The partial derivative of a function with respect to the input variables $\bm{x}_i$ is denoted with the compact notation $\partial_{x_i} f$. The total derivative of a function $f(\bm{x})$, with $f:\mathbb{R}^{n_I}\rightarrow \mathbb{R}^{n_o}$ is denoted as ${d f}/{d\bm{x}}\in\mathbb{R}^{n_i\times n_o}$ or with the short-hand notation $d_{\bm{x}} f\in\mathbb{R}^{n_i\times n_o}$. The partial derivative of vector valued functions are collected produce the \emph{Jacobian} $\mathbf{d}f/\mathbf{dx}$. The entries of the Jacobian are computed as $\mathbf{d}f/\mathbf{dx}_{i,j}=\partial_{\bm{x}_j} f_i$. In the case of a function $f:\mathbb{R}^{n_i}\rightarrow \mathbb{R}$, this becomes a row vector and is called \emph{gradient}. Many authors use the nabla symbol $\nabla f$ for the gradient, but we do not make the distinction and treat it as a Jacobian.

\textbf{Parametric and nonparametric representation}. For parametric functions $f: \bm{x}\in\mathbb{R}^{n_x}\rightarrow \mathbb{R}^{n_y}$ that depend on parameters $\bm{w}\in\mathbb{R}^{n_w}$, we use the notation $\bm{y}=f(\bm{x};\bm{w})$ or $\bm{y}=f(\bm{x}|\bm{w})$. This distinction is important for differentiating between parametric and non-parametric models. For example, we can write a parabola $y=c_2 x^2 + c_1 x + c_0$ as a parametric function $y=f(x;\bm{c})$ with $\bm{c}=[c_1,c_2,c_3]$. This notation implies that we have a script that will require in inputs both $x$ and $\bm{c}$; this is the essence of parametric modeling. However, we could put the focus on the data used for making those predictions. Assuming that the model parameters $\bm{c}$ were inferred from a training dataset $\boldsymbol{\Gamma_*}=(\bm{x}_*,\bm{y}_*)$, we might also write the parametric function as $y=f(x|\Gamma_*)$. This notation implies that our script will require input $x$ and the training data $\boldsymbol{\Gamma_*}$; this is the essence of non-parametric modelling.

\begin{figure}[htbp]
	\centering
	\includegraphics[keepaspectratio=true,width=0.8\columnwidth]
	{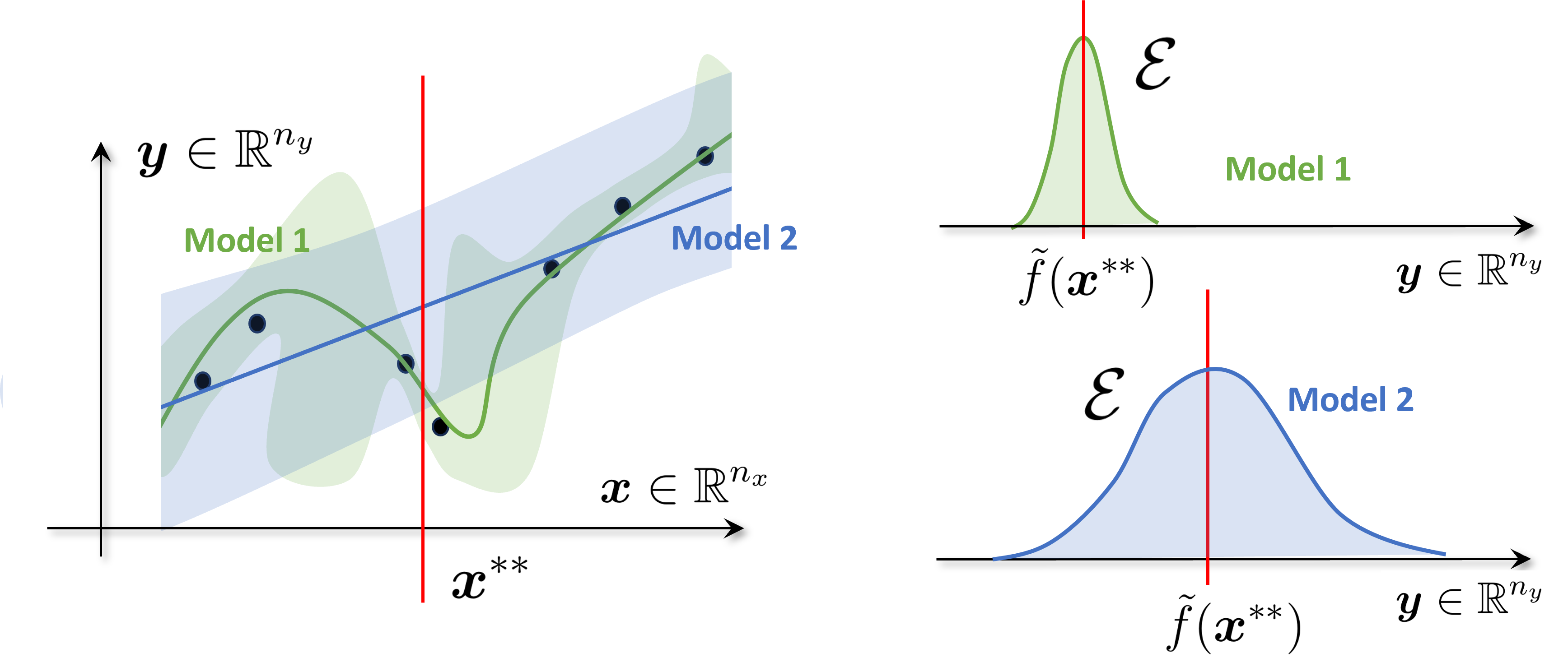}
	\caption{General overview of the regression framework, pictorially illustrated for a scalar problem. Left: Two possible models fit the training data (black dots). A prediction is requested and a stochastic process has to be fitted to the data. This can be described as in \eqref{eq1} with a deterministic model for the mean and a stochastic model for the local distribution.}\label{Fig_11}
\end{figure}

\section{General Concepts}\label{ch2_sec_1}

Regression is a subset of statistics and machine learning and, in particular, a subset of \emph{supervised} (or predictive) \emph{learning}. The goal is to learn a mapping from a continuous variable $\bm{x}\in\mathbb{R}^{n_x}$ to another continuous variable $\bm{y}\in\mathbb{R}^{n_y}$. This is in contrast to classification, where the output variable is categorical (i.e. it contains a finite set of classes/categories, such as "yes" or "no", "cat" or "dog", "laminar" or "turbulent"). Regression and classification share much of the general mathematical framework, and most of the algorithms used for one can be used for another with little to no modifications. The boundaries sometimes are so blurred that some regression methods (e.g. logistic regression) are de facto tools for classification. 

In its goal of identifying continuous functions from data, regression methods share some common grounds with other curve-fitting methods, such as interpolation or smoothing. However, the key difference is in the statistical roots: the variables $\bm{x},\bm{y}$ have a stochastic nature and thus generate a random process (that is, a distribution of possible functions). 
Our goal is to fit a random process to the data. In most cases, and indeed for all problems discussed in this lecture, we would be satisfied with a prediction of the mean function and a confidence interval around that mean. 

Let us introduce the general formalism with the help of Figure \ref{Fig_11}, which pictorially represents the problem for the case of a scalar-valued function to ease the graphical representation. Let us assume that a set of $n_*$ training points (black dots) is available. In the most general high-dimensional scenario, these could be stored in two matrices: 

\begin{equation}\label{training_data}
\bm{X}_{*} :=  \begin{bmatrix} \bm{x}_{*1}  \dots \\\bm{x}_{**2}  \dots \\ \vdots\\ \bm{x}_{n_p}\dots\end{bmatrix} \in\mathbb{R}^{n_{*} \times n_x} \quad\mbox{and}\quad \bm{Y}_{*} :=  \begin{bmatrix} \bm{y}_{*1}  \dots \\\bm{y}_{*2}  \dots \\ \vdots\\ \bm{y}_{n_p}\dots\end{bmatrix} \in\mathbb{R}^{n_{*} \times n_y} \,.
\end{equation}

To compress the notation, let us store this training data into a list $\bm{\Gamma_*}=(\bm{X}_*,\bm{Y}_*)$. The simplest model to picture a stochastic process is an additive model:

\begin{equation}\label{eq1}
\bm{Y} (\bm{X} |\bm{\Gamma_*})= \tilde{f}(\bm{X}| \bm{\Gamma_*}) + \mathcal{E}(\bm{X}| \bm{\Gamma_*})\,,
\end{equation} with $\bm{X}\in\mathbb{R}^{n_*\times n_x}$ an arbitrary set of input points and $\bm{Y}\in\mathbb{R}^{n_*\times n_y}$ the associated predictions. The first term $\tilde{f}()$ is a deterministic function that provides one output given one input. The second term $\mathcal{E}()$ is a random variable to which we associate a probability density function $\mathcal{E}$. Therefore, the first term generates a surface in $\mathbb{R}^{n_y}$ (a curve in Figure 1), and the second term generates a distribution at each input, centred on the mean prediction. That is, the stochastic term has zero average everywhere: $\mathbb{E}\{\mathcal{E}(\bm{X})\}=0$, with $\mathbb{E}$ the expectation operator. We usually use the deterministic model $\tilde{f}$ to make predictions and the probabilistic model $\mathcal{E}$ to provide uncertainties associated with these predictions.  

To make the discussion more concrete, we consider two models, indicated in blue and green in Figure 1. A prediction is required for both of them in a new point $\bm{x}_{**}$. The figure on the right shows the distributions the two models generate at new locations. The width of the distribution at each location is linked to the width of the shaded areas on the right-hand side. Model 1 is much more complex and has a large variability in the distribution width compared to model 2. Which one of the two is most appropriate? We will never know, but we have the tools to make an educated guess.

\subsection{A probabilistic perspective: The MLE}\label{ch_2_sec_1_p_1}

The \emph{maximum likelihood estimation} (MLE) is an essential principle for fitting a stochastic process. 
We here introduce it in the simplest possible setting and refer the reader to \cite{bishop2006pattern,Deisenroth,AbuMostafa2012book,murphy2012machine,Watt} for a more extensive treatment. 

First, we consider the regression of a univariate (scalar) problem, that is $n_x=n_y=1$ and $\tilde{f}:x\in\mathbb {R}\rightarrow y \in\mathbb{R}$. Second, we consider the simplest assumption for the stochastic contribution: we assume that all distributions we have seen in Figure \ref{Fig_11} are Gaussian in every $x$ and have all the same standard deviation $\sigma_y\in\mathbb{R}^+$. This assumption leads to the well-known \emph{least square problem}. Different assumptions on the stochastic contributions leads to different minimization problems.

As a model for the mean prediction, we take a generic parametric function $y=\tilde{f}(x;\bm{w})$. This function could be a polynomial, a radial basis function expansion, an artificial neural network to mention some classic examples. Equipped with a model for the mean prediction and a model for the stochastic contribution in \eqref{eq1}, we can now calculate the probability of observing a certain outcome $y = y_*$ for a specific input $x=x_*$. In the aforementioned setting, this reads 

\begin{equation}
p(y = y_*)\propto \exp{\biggl( -\frac{[y_*-\tilde{f}(x_*;\bm{w})]^2}{2 \sigma^2_y}}\biggr)\,.
\end{equation}

Given the $n_*$ sample points in the training set $\Gamma_*=(\bm{x}_*,\bm{y}_*)$, we shall now ask ourselves: \emph{What is the likelihood of observing the collected data if our model is valid?} The fact that this specific set of data has been collected (and not others) suggests that this specific set has a much higher probability of occurring than others\footnote{It is important to remember that in a probabilistic framework, no outcome can be deemed impossible. As illustrated in Figure \ref{Fig_11}, all curves within the shaded area have a high probability of accurately representing the data, indicating their validity. In contrast, curves consistently falling outside this area are less likely to be representative. However, no curve is entirely impossible.}

According to the MLE principle, fitting a model means looking for the model that best explains the collected data, i.e. the model according to which the probability of observing that specific dataset is the highest. In other words, defining the \emph{likelihood} as $p(\bm{\Gamma}_*| \bm{Y}(\bm{X}))$ the probability of observing the data $\bm{\Gamma}_*$ if the model $\bm{Y}(\bm{X})$ is ``true'', we seek the model that maximizes the likelihood.
The assumption of uniform $\sigma_y$ in each point of the domain implies that the outcome at each location is entirely independent of the outcome in other locations. The same must be true for the data we have collected, which thus are independent and identically distributed (i.i.d.). Therefore, the likelihood of observing the specific sequence we have collected, according to our current model, is 

\begin{equation}
\begin{split}
\label{likelihood}
p(\bm{y}=\bm{y}_*|\bm{w})&\propto \prod^{n_p-1}_{i=0} \exp{\biggl(- \frac{(\bm{y}_{*i}-\tilde{f}(\bm{x}_{*i};\bm{w}))^2}{2 \sigma^2_y}}\biggr)\\
& =\exp{\biggl(- \sum^{n_p}_{i=0}\frac{(\bm{y}_{*i}-\tilde{f}(\bm{x}_{*i};\bm{w}))^2}{2 \sigma^2_y}}\biggr)\\
& =\exp{\biggl(- \frac{||\bm{y}_*-\tilde{f}(\bm{x}_*;\bm{w})||_2^2}{2 \sigma^2_y}}\biggr) \,.
\end{split}
\end{equation} having used basic properties of the exponential and having introduced the $l_2$ norm $||\bullet||_2$. Without necessarily involving logarithms and calculus\footnote{To find the maximum of \eqref{likelihood}, one usually takes the logarithm on both sides to obtain the log-likelihood and proceeds to show that the MLE requires the minimization of the MSE (e.g. \citet{bishop2006pattern}).}, it is intuitive that maximizing the exponential (hence the likelihood) requires minimizing its argument.  Hence, maximizing the likelihood is equivalent, in this case, to minimizing the mean square error (often referred to as MSE), which reads: 

\begin{equation}
\label{MSE}
\mathcal{J}(\bm{w})=\frac{1}{n_*}\sum^{n_p-1}_{i=0}\bigl[\bm{y}_{*i}-\tilde{f}(\bm{x}_{*i};\bm{w})\bigr]^2=\frac{1}{n_*}||\bm{y}_*-\tilde{f}(\bm{x}_*;\bm{w})||_2^2\,.
\end{equation}

It is also interesting to note that, in this case, the MSE corresponds to\footnote{See \cite{taboga} for a detailed derivation.} the sample variance of the distribution $\mathcal{E}$. Hence, given $\bm{w}^{*}:=\argmin_w \mathcal{J}(\bm{w})$, one can estimate $\sigma^2_y\approx \mathcal{J}(\bm{w}_*)$.

The best method to minimize \eqref{MSE} depends on the kind of parametric function. A closed-form solution is available for linear methods (e.g., radial basis function regression), while numerical optimization is required for nonlinear methods (e.g., artificial neural networks).

The generalization to higher dimensions is straightforward. The MSE is the most popular cost function for training machine learning algorithms, but many alternatives exist; we discuss these briefly in Section \ref{ch_2_sec_1_p_3}. First, let us return to the problem of fitting a random process to the data and evaluate its predictions.

% f(\bm{x};\bm{w})

\subsection{Bootstrapping and cross-validation }\label{ch_2_sec_1_p_2}

Let us assume that the simplified stochastic model of Gaussian with uniform variance $\sigma^2_y$ is appropriate. We have identified a suitable parametric model $y=\tilde{f}(x;\bm{w})$ and the optimal set of parameters $\bm{w}_*$ that minimize the MSE in \eqref{MSE}. Approximating $\sigma^2_y\approx$ MSE, we can now draw the mean prediction $f(x;\bm{w}_*)$ and identify a constant shaded area around it. For example, for the canonical confidence interval of $95\%$, we would draw the boundaries of the shaded area as $f(x;\bm{w}_*)\pm 1.96 \sigma_y$. The random process we infer would be:

\begin{equation}
\label{infty}
\bm{Y}(\bm{X}|\bm{w}_*)=\tilde{f}(x|\bm{w}_*)+\mathcal{N}(0,\mathcal{J}(\bm{w}_*))\,,
\end{equation} where $\mathcal{N}(0,\mathcal{J}(\bm{w}_*)$ is a Gaussian with zero mean and covariance $\mathcal{J}(\bm{w}_*)$.

Are we done? Not quite. The model in \eqref{infty} is valid only in the theoretical limit of infinitely many training data points. The parameters $\bm{w}*$ we have identified are optimal (in the sense of minimizing the MSE) only for the specific training dataset $\Gamma=(\bm{x}_,\bm{y}_*)$. But what guarantees do we have that these parameters will perform well on new, unseen data? In general, we have no such guarantee. We can only hope that, if the training dataset is sufficiently large and representative, the resulting parameter estimate will generalize reasonably well.

A classic method for addressing the problem is statistical resampling. This usually takes the forms of \emph{bootstrapping} or \emph{cross-validation} (see \cite{Kim2009}). Both methods seek to reproduce multiple ensembles of training data from one single dataset, to assess the performance of a predictive model in datasets that were not used during the training. Although the notion of Bootstrapping arises in a more general statistical context (see \citet{Efron1993,Davidson2009} for an extensive overview), we here solely focus on the problem of model assessment. 

In both cases, the available data is split into \emph{training data} to evaluate the \emph{in-sample error}\footnote{which we can use to estimate $\sigma_y$, in the simplest stochastic model considered in the previous section.} and \emph{testing data} to evaluate the \emph{out-of-sample error} and assess how well the model generalizes.

Bootstrapping and cross-validation differ in how they make use of the available data. In \emph{bootstrapping}, the dataset is sampled randomly \emph{with replacement}, meaning that data points may be selected multiple times or not at all. This approach is generally appropriate when a large dataset is available. For instance, if $n_p = 1000$ data points are given, one may train the model $n_E = 100$ times, each time selecting $n_* = 700$ points for training and using the remaining $n_{**} = 300$ for testing. As $n_*$ increases toward $n_p$, the likelihood that the same data points appear in multiple ensembles increases, which reduces variability across the ensemble. Since sampling is performed with replacement, duplicated entries may also occur within a single training set. Notably, one could even choose $n_* = n_p$ and still obtain slightly different training sets across ensembles.

In \emph{cross-validation}, by contrast, the data is partitioned into $K$ disjoint folds, and \emph{no} data point appears in more than one fold. For $K = 10$, the model is trained $n_E = 10$ times, each time using $n_* = 900$ data points for training and $n_{**} = 100$ for testing: at each iteration, one fold is used for testing and the remaining folds for training. The limiting case is the \emph{leave-one-out} strategy, where $K = n_p$, so the model is trained $n_E = n_p$ times, each time with $n_* = n_p - 1$ and $n_{**} = 1$.

\begin{figure}[htbp]
	\centering
	\includegraphics[keepaspectratio=true,width=0.8\columnwidth]
	{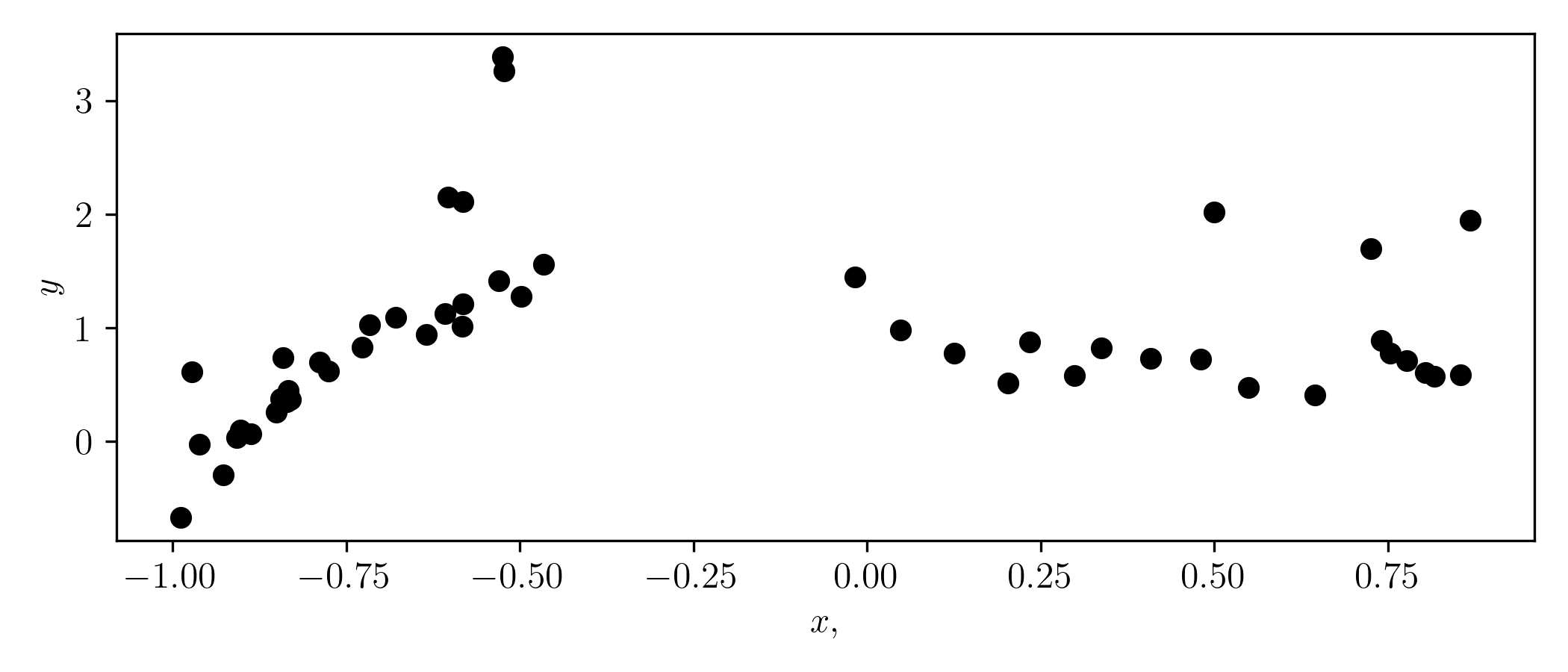}
	\caption{Dataset for tutorial 1, to illustrate the usage of cross-validation }\label{Fig2}
\end{figure}

Cross-validation is usually preferred for model validation, while bootstrapping is usually preferred for uncertainty quantification using a process called \emph{bagging} (short for Bootstrap Aggregating, proposed by \cite{Breiman1996}). We illustrate their usage with our first Python exercise. We consider the dataset in Figure \ref{Fig2}. This consists of a small dataset ($n_p=60$) with a significant noise level and several outliers. Moreover, data is lacking in a critical region. We test two polynomial models of different degrees and evaluate their performances.

The following Python function takes in input the available data ({x,y}), the order of the polynomial we want to test ({n\_O}), the number of ensemble members we will re-sample ({n\_E}) and the $\%$ of data we seek to keep as testing ({tp}) at each time.

\begin{centering}
	\begin{lstlisting}[language=Python,linewidth=13.8cm,xleftmargin=.05\textwidth,xrightmargin=.05\textwidth,backgroundcolor=\color{yellow!10}]
def Ensemble_Train_poly(x, y, n_O, n_E=100, tp=0.3):    
 '''
 see python file for the documentation     
 '''
 # in sample error of the population
 J_i = np.zeros(n_E) 
 # out of sample error of the population
 J_o = np.zeros(n_E) 
 # Distribution of weights
 w_e = np.zeros((n_O+1, n_E))     
 for j in range(n_E):
  # Split the dataset into training and testing    
  x_s, x_ss, y_s, y_ss = train_test_split(x, y, 
                         test_size=tp)   
  # Fit the polynomial on the training data
  w_s = np.polyfit(x_s, y_s, n_O)
  # store the model parameters
  w_e[:,j] = w_s
  #-------- in-sample performance ---------
  # Make a prediction on the training data
  y_tilde_s = np.polyval(w_s, x_s)
  # In-sample error
  J_i[j] = 1/len(x_s) * np.linalg.norm(y_tilde_s-y_s)**2
  #-------- out-of-sample performance ---------
  y_tilde_ss = np.poly(w_s, x_ss)
  # Out of sample error
  J_o[j] = 1/len(x_ss) * np.linalg.norm(y_tilde_ss-
                        y_ss)**2  
 return J_i, J_o, w_e
	\end{lstlisting}
\end{centering}

The code returns the in-sample and out-of-sample MSE \eqref{MSE} ({J\_i, J\_o}) in each of the resampled ensembles and the population of parameters as a matrix ({w\_e}). The function repeats $n_E$ times the following steps: (1) randomly split the data into training and testing portions (line 10), (2) fits the model minimizing the MSE (line 12), (3) makes predictions on the training data and evaluates in-sample performances (lines 17-19), (4) makes predictions on the testing data and evaluates out-of-sample performances (lines 21-23).

\begin{figure}[htbp]
     	\centering  
\includegraphics[width=12cm]{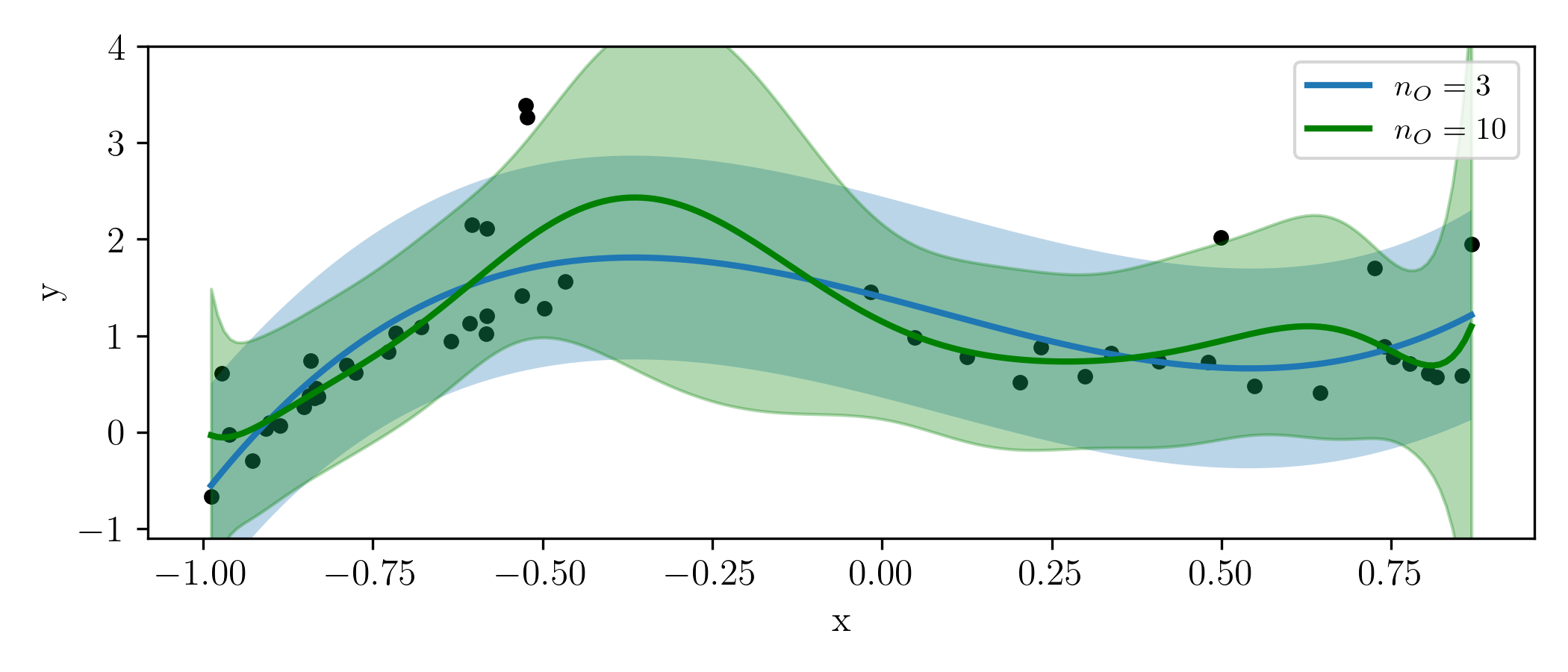}\\
\includegraphics[width=6cm]{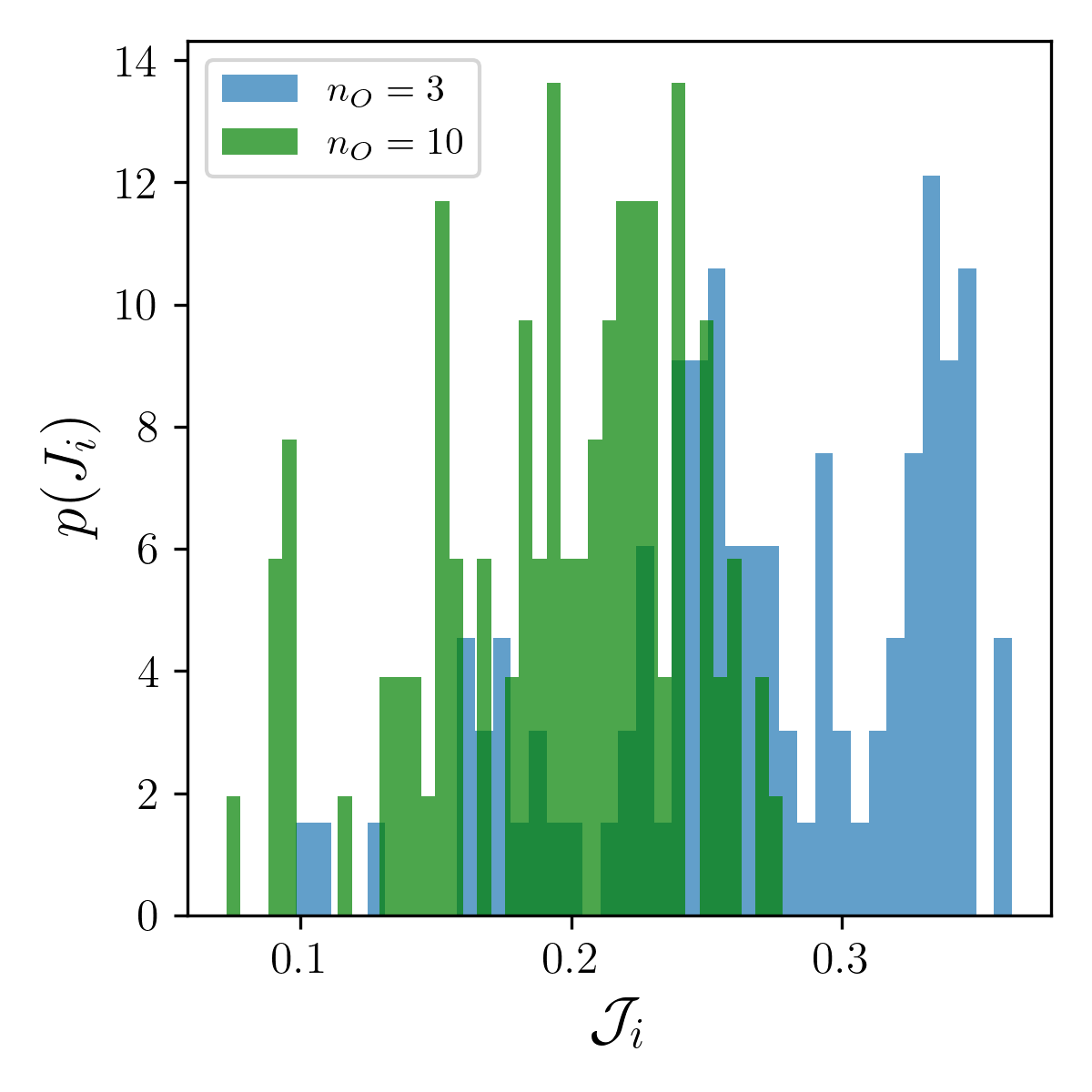}		
\includegraphics[width=6cm ]{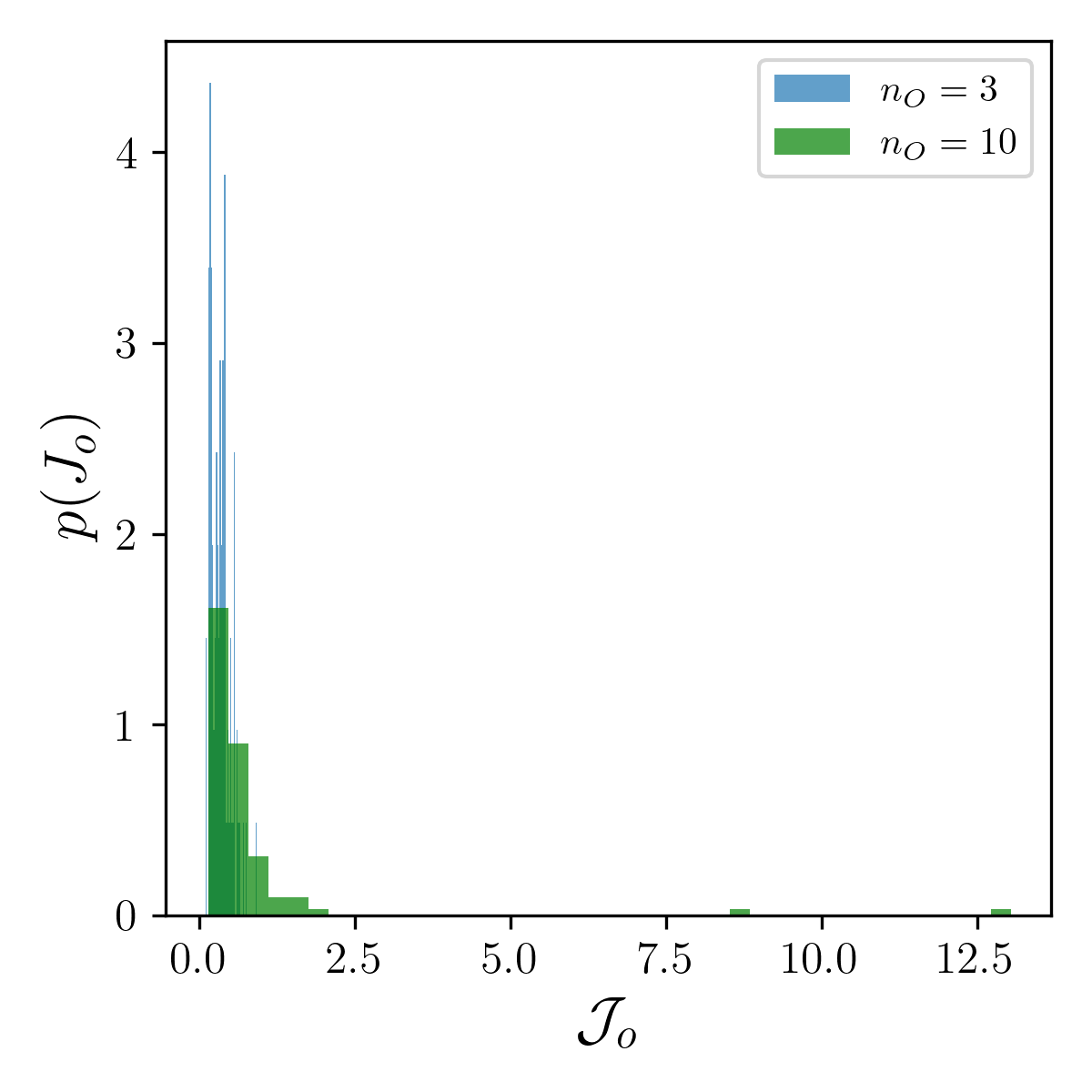}    
\caption{Tutorial to illustrate the usage of bootstrapping to estimate uncertainties via bootstrapping. Top: prediction of the two models versus data. Bottom: distribution of in-sample MSE (left) and out-of-sample MSE (right). The more complex model is more prone to overfitting.} 
     	\label{Fig3}
\end{figure}

We do not explore the details of MSE minimization on line 12. It is enough to recall that a closed-form solution exists for a polynomial model, requiring only the solution of a linear system—handled by the polyfill function. 

Since the process returns n\_E models, we could test them all and use the results to make a plot like the one sketched in Figure \ref{Fig_11} with the following Python function:

\begin{centering}
	\begin{lstlisting}[language=Python,linewidth=13.8cm,xleftmargin=.05\textwidth,xrightmargin=.05\textwidth,backgroundcolor=\color{yellow!10}]
def Ensemble_Pred_poly(xg, w_e, J_i_mean):
    '''see python file for the documentation'''
    # Get n_p, n_O and n_e
    n_p = len(xg); n_Op1, n_e=np.shape(w_e)
    # prepare the population of predictions in y:
    y_pop = np.zeros((n_p, n_e))    
    for j in range(n_e):   # loop over the ensemble  
        # predict for each set of w
        y_pop[:,j] = Poly_model_Pred(xg, w_e[:,j])
    # The mean prediction will be:
    y_e = np.mean(y_pop, axis=1)   
    # the ensamble std:
    Var_Y_model = np.std(y_pop, axis=1)**2    
    # Compute the final uncertainty:
    Unc_y = np.sqrt(J_i_mean + Var_Y_model)    
    return y_e, Unc_y
	\end{lstlisting}
\end{centering}

The output gives the mean prediction functions and the width of the probability density function sitting on each of these. Note that the final uncertainty in line 16 is the sum of two contributions, assuming that these are independent. The first is the estimated variance $\sigma^2_y$ from the data and is computed from the in-sample error. The second is the variance of the predictions due to the sensitivity of the model to small variations in the dataset. The first contribution is large when the model is \emph{underfitting}. This means that the model is too simple to explain the data. The second contribution is large when the model is \emph{overfitting}. This means that the model is too complex for the data at hand and thus becomes too sensitive to minor changes in the data. Increasing the model complexity generally reduces the first contribution but increases the second. The best model is the one that offers the best compromise. The less data we have and the more complex a model is, the higher the risk of overfitting.

Let us compare the performances of two models: the first with n\_O=3 and the second with n\_O=10. The results are shown in Figure \ref{Fig3}, and the reader is referred to the provided Python files to reproduce the results. 

The predictions of both models seem reasonable, with the model with higher order appearing 'wiggly'. The most complex model has a generally lower in-sample error (\verb|J_i|) and higher out-of-sample error (\verb|J_o|). The distribution of out-sample error is particularly skewed, with some ensemble producing a particularly high MSE. This is a classic footprint of overfitting. It is interesting to note that the width of the uncertainty region is almost constant for the simplest model but oscillates considerably for the most complex, especially in the regions lacking data. This is a second footprint of overfitting, as the uncertainty is mainly dominated by the model error, which has a strong variability. This is also larger in the area where data is missing.

A faster evaluation of the model performance can be carried out using cross-validation without looking at the uncertainty distribution but only considering the out-of-sample MSE. Here's a Python function to compute the cross-validation score of a polynomial model.

\begin{centering}
	\begin{lstlisting}[language=Python,linewidth=13.8cm,xleftmargin=.05\textwidth,xrightmargin=.05\textwidth,backgroundcolor=\color{yellow!10}]
def CV_poly_fold(x_s, Folds, n_O):
 '''refer to the python files for the docs'''
 # This prepares the splitting
 kf = KFold(n_splits=Folds, shuffle=True, 
            random_state=42)
 mse_list = [] # prepare the list 
 # This runs the splitting
 for train_index, test_index in kf.split(x_s):
   # split the folds   
   x_train,y_train = x_s[train_index],
                       y_s[train_index]
   x_test,y_test = x_s[test_index], y_s[test_index]
   # we train on training folds:
   w_s = np.polyfit(x_train, y_train, n_O)
   #predict on testing folds
   y_tilde_ss = np.polyval(w_s, x_test)    
   # compute MSE in test:
   J_o = 1/len(x_test) * np.linalg.norm(y_tilde_ss-
            y_test)**2
   #store out-of-sample mse
   mse_list.append(J_o)
         
 average_mse_score = np.mean(mse_list) 
 std_mse_score = np.std(mse_list)      
 print(f'Average MSE score for n_O={n_O}:
        {average_mse_score}') 
 print(f'STD of MSE score for n_O={n_O}: 
        {std_mse_score}')
 return mse_list
	\end{lstlisting}
\end{centering}

This script leverages the function KFOLD from scikitlearn for the splitting. Using 5 folds, this script returns an average MSE core of 0.33 for $n_O=3$ and 0.46 for $n_O=10$. The standard deviation on the MSE is 0.16 for $n_O=3$ and 0.22 for $n_O=10$: the most complex model performs worse by all metrics. The Bayesian framework and the kernel formalism allow us to bypass the need for bootstrapping in the uncertainty estimation \citep{mendezHandsOnML}.

\subsection{More Cost Functions}\label{ch_2_sec_1_p_3}

We learned that the MSE cost function can be derived from the MLE under the hypothesis that the stochastic contribution of our model is Gaussian, independent and identically distributed with a constant variance. Releasing the assumption of constant variance and assuming that this can vary within the domain\footnote{This variability is called heteroscedasticity, as opposed to the homoscedastic (constant variance) from the previous test case}, the same procedure would take us to a weighted norm of the kind

\begin{align}
\label{MSE_w}
\mathcal{J}(\bm{w}) 
&= \frac{1}{n_p}\bigl(\bm{y}_{*}-\tilde{f}(\bm{x};\bm{w})\bigr) \bm{\Sigma}^{-1}\bigl(\bm{y}_{*}-\tilde{f}(\bm{x};\bm{w})\bigr) \notag\\
&= \frac{1}{n_p} \left\| \bm{y}_{*}-\tilde{f}(\bm{x}_i;\bm{w}) \right\|_{\bm{\Sigma}}^2\,.
\end{align} where the matrix $\bm{\Sigma}$ is the covariance of the stochastic contribution. We still assume that the distributions in each location are independent. Cost functions with weighted norms of this kind are popular in data assimilation, for example, in the formulation of the Kalman filter (see \citealt{Asch2016,Bocquet2023,Bocquet2011} for an overview).

However, the zoology of regression cost functions is vast (see \cite{Hastie2009}) and is mainly promoted by the need to handle outliers (see \citet{Andersen2007}), to which all quadratic losses (weighted or not) are overly sensitive. A cost function that makes the regression less influenced by outliers is the $l_1$ penalty, obtained by replacing the $l_2$ norm in \eqref{MSE} with an $l_1$ norm. This cost function can be derived assuming that the stochastic contribution follows a Laplacian distribution \citep{Nair2022}.

Variants to the $l_1$ cost that are particularly robust against outliers are piece-wise formulations, such as, for example, the Huber loss function \citep{Huber1964}. Defined as $\bm{e}_i=\bm{y}_i-f(\bm{x}_i;\bm{w})$ the error in a prediction for a parametric model, the Huber loss reads

\begin{equation}
\label{Huber}
\mathcal{J}(\bm{w})=\frac{1}{n_p}\sum^{n_p-1}_{i=0} L_{\delta}(\bm{e};\delta)\, \quad\mbox{with}\quad  L_{\delta}(\bm{e};\delta) =
\begin{cases} \frac{1}{2}\bm{e_i}^2 & \mbox{for}\, |\bm{e_i}| \leq \delta
\\  \delta\bigl(|\bm{e_i}|-\frac{1}{2}\delta\bigr)& \mbox{otherwise}\end{cases}  \,.
\end{equation}

This cost function results from the convolution of the absolute value function with the rectangular function, scaled and translated appropriately. It basically "blends smoothly" the $l_2$ loss for minor errors (smaller than $\delta$) with the $l_1$ loss for significant errors. 

A variant commonly used in Support Vector Regression (SVR, see \cite{Smola2004sac}) uses the idea of $\epsilon$ sensitiveness and reads:

\begin{equation}
\label{E_epsilon}
\mathcal{J}(\bm{w})=\frac{1}{n_p}\sum^{n_p-1}_{i=0} E_{\epsilon}(\bm{e};\epsilon)\, \quad\mbox{with}\quad  E_{\epsilon}(\bm{e_i};\epsilon) =
\begin{cases} 0 & \mbox{for}\, |\bm{e_i}| \leq \epsilon\\
|\bm{e_i}|-\epsilon & \mbox{for}\, |\bm{e_i}| > \epsilon
\end{cases}  \,.
\end{equation}

That is, the loss is zero for data sets within a $\pm \epsilon$ region around the predictions. Finally, the last class of cost functions that we shall briefly touch on in these notes is the one of \emph{penalized} regression. These cost functions add term (penalty) for parameters. The most classic approach (also known as Ridge regression; see \cite{Wieringen2015}) adds the $l_2$ norm of the weights to \eqref{MSE}:

\begin{equation}
\label{MSE_Ridge}
\mathcal{J}(\bm{w})=\frac{1}{n_p}||\bm{y}_{*i}-\tilde{f}(\bm{x}_i;\bm{w})||_2^2\, + \alpha ||\bm{w}||^2_2\,,
\end{equation} where $\alpha\in\mathbb{R}^+$ is an user defined parameter. The scope of a $l_2$ penalty is to increase the robustness of the regression against overfitting. 
On the other hand, the LASSO (Least Absolute Shrinkage and Selection Operaton) regression by \cite{LASSO} uses an $l_1$ penalty:

\begin{equation}
\label{Lasso_R}
\mathcal{J}(\bm{w})=\frac{1}{n_p}||\bm{y}_{*i}-\tilde{f}(\bm{x}_i;\bm{w})||_2^2\, + \alpha ||\bm{w}||_1\,.
\end{equation}

The scope of a $l_1$ penalty is to promote sparsity, that is, a model in which many of the entries in the parameter vector $\bm{w}$ are null.

\subsection{Methods for parametric regression}\label{ch_2_sec_2_1}

Parametric methods can be broadly classified into linear and non-linear methods depending on whether their predictions are linearly related to the parameters or not.

\paragraph{Linear Methods.} The polynomial regression considered in the previous section is an example of a linear method, in the sense that the model is linear with respect to the parameters $\bm{w}$. For instance, in the case of a cubic polynomial model $\tilde{f}: \mathbb{R}\rightarrow\mathbb{R}$, predictions at new input points $\bm{x}_{**}\in\mathbb{R}^{n_{**}}$ can be written as the following matrix–vector product:
\begin{equation}
\label{poly}
\bm{y}_{**}
=
\begin{bmatrix}
|  \\
\bm{y}_{**}\\
|
\end{bmatrix}
=
\begin{bmatrix}
| & | & | & | \\
\bm{x}_{**}^3 & \bm{x}_{**}^2 & \bm{x}_{**} & \bm{1} \\
| & | & | & | 
\end{bmatrix}
\begin{bmatrix}
\bm{w}_0 \\
\bm{w}_1 \\
\bm{w}_2 \\
\bm{w}_3
\end{bmatrix}
= \bm{\Phi}(\bm{x}_{**})\,\bm{w}\,,
\end{equation}
where $\bm{\Phi}(\bm{x}_{**}) \in \mathbb{R}^{n_{**}\times 4}$ is the \emph{feature matrix} for the cubic polynomial regression, evaluated at the new points $\bm{x}_{**}$. The prediction is therefore a linear combination of the columns of $\bm{\Phi}(\bm{x}_{**})$, which act as a \emph{basis}. In cubic polynomial regression this basis is generated by the functions $\{1, x, x^2, x^3\}$, but other sets of basis functions can be used (see, e.g., \cite{bishop2006pattern}) without changing the structure of training or prediction. More generally, for a linear parametric model $\tilde{f}:\mathbb{R}\rightarrow\mathbb{R}$, we can write
\begin{equation}
\label{linear_parametric}
\bm{y}(\bm{x})
=
\sum_{r=0}^{n_b-1} \phi_r(\bm{x})\,w_r
=
\bm{\Phi}(\bm{x})\,\bm{w}\,,
\end{equation}
where $\bm{\Phi}(\bm{x})\in\mathbb{R}^{n_s\times n_b}$ collects the basis functions $\phi_r(\bm{x})$ as its columns.

\begin{comment}
% introduce it in the cost function 
$$\mathcal{J}(\bm{w})= ||\bm{y}_*-\bm{\Phi}(\bm{x}_*) \bm{w}||^2_2$$

The you can show that the gradient is

$$\frac{d \mathcal{J}}{d \bm{w}}=2\bm{\Phi}^T(\bm{x}_*) (\bm{y}_*-\bm{\Phi}(\bm{x}_*) \bm{w})$$

Setting it equal to zero we get

$$\Bigl [\bm{\Phi}^T(\bm{x}_*)\bm{\Phi}(\bm{x}_*)\Bigr ]\bm{w}=\bm{\Phi}^T\bm{y}_*$$

$$\frac{d \mathcal{J}}{d \bm{w}}=\bm{0}$$

$$\bm{\Phi}^T(\bm{x}_*)\bm{\Phi}(\bm{x}_*) \succ 0$$

\end{comment}

In the case of Radial Basis Function regressions, the basis functions are \emph{radial}, in the sense that they solely depend on the distance from the points in which they are \emph{collocated}. In first tutorial exercise in Chapter 3, we use Gaussian RBFs defined as 

\begin{equation}
    \varphi_r(\bm{x}\vert\bm{x}_{c,k}, c_k) = \text{exp}\left( -c_k^2 \vert\vert \bm{x} - \bm{x}_{c,k} \vert\vert^2 \right),
    \label{eq:gaussian_rbfs}
\end{equation} where $\bm{x}=(x,y,z)$ is the coordinate vector where the basis is evaluated in a 3D domain, $\bm{x}_{c,k}=(x_{c,k},y_{c,k},z_{c,k})$ is the k-th collocation point and $c_k$ the shape parameter of the basis, defining its width. The definition of the basis in 3D therefore requires defining the matrix $\bm{X}_c\in\mathbb{R}^{n_b\times 3}$ collecting the collocation points of all bases and the vector $\bm{c}\in\mathbb{R}^{n_b}$ collecting their shape factors.

Linearity makes the training of these models particularly simple. In the case of classic quadratic cost functions such as the MSE in \eqref{MSE}, its weighted \eqref{MSE_w} or $l_2$ penalized \eqref{MSE_Ridge} version, an analytic solution for the optimal set of parameters can be derived. In the Ridge regression, that is, the minimization of  \eqref{MSE_Ridge} for a linear parametric model like in \eqref{poly}, we have:

\begin{equation}
\label{ridge}
\bm{w}_*=\bigl(\bm{\Phi}^T(\bm{x}_{*})\bm{\Phi}(\bm{x}_{*})+\alpha\bm{I}\bigr)^{-1}\bm{\Phi}^T(\bm{x}_{*})\bm{y}_*\in\mathbb{R}^{n_b}\,,
\end{equation} where $\bm{I}\in{\mathbb{R}^{n_b\times n_b}}$ is the identity matrix and $n_b$ the number of bases used by the model. As in all parametric methods, the training data is no longer needed to make predictions once the parameters are available. The same formulation extends directly to vector-valued outputs $\tilde{f}:\mathbb{R}^{n_x}\rightarrow\mathbb{R}^{n_y}$ by keeping the feature matrix $\bm{\Phi}$ unchanged and letting the parameter vector become a weight matrix $\bm{W}\in\mathbb{R}^{n_b\times n_y}$, so that $\bm{Y}=\bm{\Phi}\bm{W}$. In this case, each column of $\bm{W}$ corresponds to one output component, all sharing the same basis representation.

In the case of more sophisticated cost functions, for which an analytic solution such as \eqref{ridge} does not exist, numerical optimization is required. This usually takes the form of a gradient-based approach since the gradient of the model prediction with respect to the parameters is simply the feature matrix itself $d\tilde{f}/d\bm{w}(\bm{x})= \bm{\Phi}(\bm{x})$. Finally, linear methods are particularly interesting for uncertainty quantification: The bootstrapping approach could be significantly accelerated leveraging the model linearity. More specifically, after the ensemble training and after having derived a population of possible weights, statistics on the weight distribution could be inferred and propagated to the prediction of the ensemble \emph{without interrogating all the models} \citep{mendezHandsOnML}.  However, the effectiveness of linear methods is highly dependent on the choice of the basis. These methods should always be prioritized in relatively low-dimensional problems and when a careful craft of the basis is feasible. Nonlinear methods can have larger model capacity (i.e., be able to represent more complex functions) while requiring less engineering in the model formulation, but they are generally much more difficult to train. The most natural non-linear models are artificial neural networks (ANNs). We briefly review these in the following.

\paragraph{Artificial Neural Networks (ANNs).} Artificial neural networks (ANNs) are among the most widely used nonlinear modeling tools in machine learning. Originally introduced in the late 1950s as simplified abstractions of the human brain, their biological analogy is now largely historical and not essential for our purposes. In the early developments of ANNs, this analogy has triggered scientific controversies and exaggerated (and unfulfilled\footnote{Here's an excerpt from an article in the \textsc{New York Times} from 8 July 1958: \emph{The Navy revealed the embryo of an electronic computer today that it expects will be able to walk, talk, see, write, reproduce itself and be conscious of its existence.} Not happening. Right?}) claims (see \cite{Olazaran1993}) that resulted in skepticism and a drop in research interest and funding. Today, regained interest in ANNs is fueled by a tremendous increase in computer power (particularly recent developments in GPU technology), the availability of data, improvements in training algorithms, and the diffusion of powerful and accessible open-source libraries such as \textit{Tensorflow}\footnote{See \url{https://www.tensorflow.org/}} or \textit{Pytorch}\footnote{See \url{https://pytorch.org/}}. The relevance of ANNs research has given them their own subfield of machine learning, called {\bf Deep Learning} (with the adjective ``deep'' referring to networks with many layers).

ANNs are distributed architectures with many simple connected units (called neurons) organized in layers \citep{Goodfellow2016book}. The mapping from input to output takes the recursive form:

\begin{equation}
	\label{ANN}
	\bm{y}=\tilde{f}(\bm{x};\bm{w})=a^{(L)}\left(\bm{z}^{(L)}\right)\quad \mbox{with}\quad\begin{cases}
		\bm{y}^{(1)}=\bm{x}, \,\bm{y}^{(l)}=a^{(l)}\left (\bm{z}^{(l)}\right) & \\
		\bm{z}^{(l)}=\bm{W}^{(l-1)}\bm{y}^{(l-1)}+\bm{b}^{(l)} & 
	\end{cases} \,,
\end{equation} with $\quad \mbox{and}\quad l=2,3,\dots L$. Here $\bm{y}^{(l)},\bm{z}^{(l)},\bm{b}^{(l)}\in\mathbb{R}^{n_l\times 1}$ are the \emph{output, activation} and \emph{bias vector} of the layer $l$, composed of $n_l$ neurons, $a^{(l)}$ is the activation function in each layer, $\bm{W}^{(l-1)}\in\mathbb{R}^{n_l\times n_{l-1}}$ is the matrix that contains the weights connecting the layer $l-1$ with the layer $l$, and $L$ is the number of layers. The vector $\bm{w}\in\mathbb{R}^{n_w\times 1}$ collects all the weights and biases across the network, hence $n_w=n_L n_{L-1}+n_{L-2} n_{L-3}+\dots n_2 n_1 + n_L+n_{L-1}+\dots n_2$ in the case of a fully connected feed-forward network. The input layer of the ANN consists of $n_x$ neurons that feed the input directly into the second layer, that is, no activation function or biases are applied in the first layer.

The activation functions are nonlinear functions such as hyperbolic tangents, sigmoids, or piecewise-defined functions like the Exponential Linear Unit (ELU) and its variants (see \cite{Goodfellow2016book,prince2023understanding,bishop2023deep}). These functions are what introduce nonlinearity into the model: if linear activations were used in every layer, the recursive composition would collapse to a single matrix multiplication, and the resulting model would remain linear.

\begin{figure}[htpb]
\center
 \includegraphics[width=0.8\textwidth]{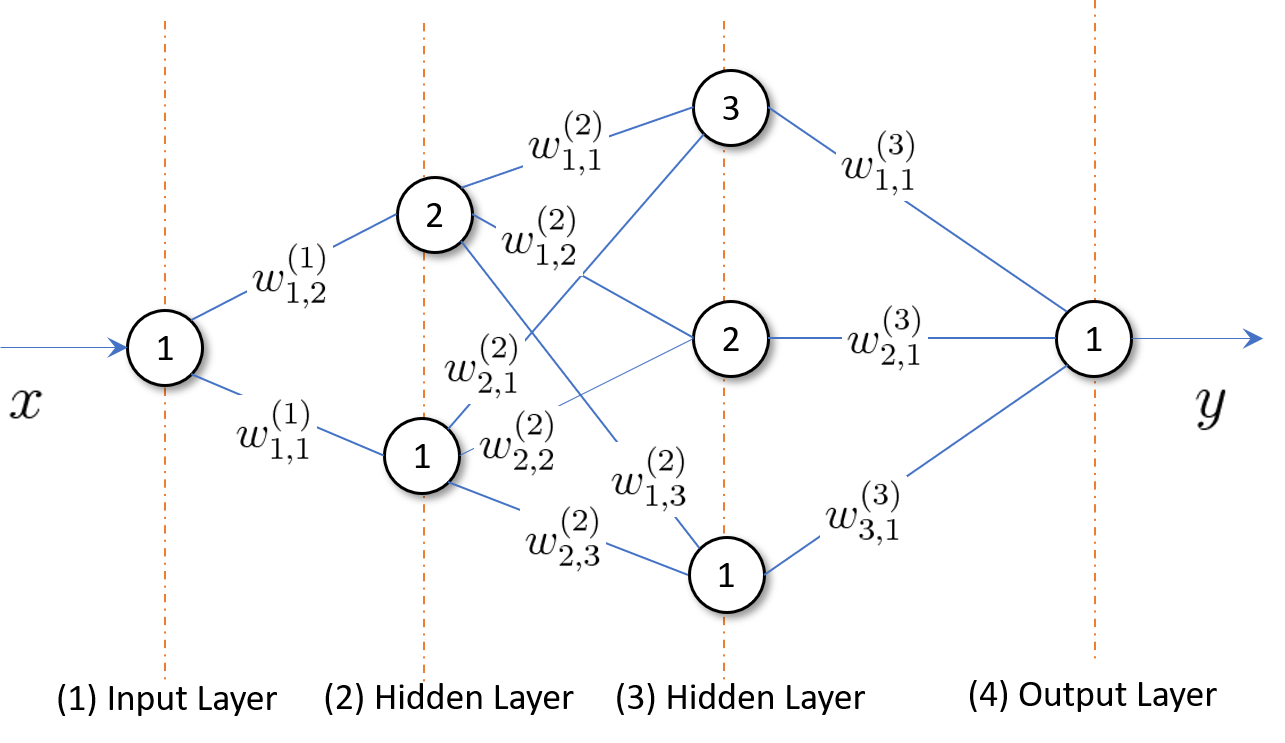}
    \centering
    \caption{A simple example of feedforward, fully connected architecture with two hidden layers and a total of seven neurons.}
    \label{ANN_example}
\end{figure}

To illustrate the recursive architecture of an ANN, consider the simple example in Figure~\ref{ANN_example}. The network consists of seven neurons arranged in four layers: one input layer, one output layer, and two intermediate \textbf{hidden layers}. As a parametric model $\tilde{f}:\mathbb{R}\rightarrow\mathbb{R}$, it contains a single neuron in both the input and output layers. Neurons are numbered from top to bottom. The network is \textbf{fully connected}, meaning each neuron in one layer is connected to all neurons in the next layer, and \textbf{feed-forward}, meaning information flows strictly from one layer to the next. Feed-forward networks are often called \textbf{Multilayer Perceptrons} (MLPs), a historical reference to the original \emph{Perceptron}, a single-neuron model developed for binary classification \citep{Minsky}.

A feedforward network is a {\bf static model} that maps a set of inputs to outputs independently from one another. It is thus a memoryless model. Conversely, {\bf recurrent neural networks} are {\bf dynamical systems} characterized by feedback connections (e.g. from output to input) so that each input triggers a sequence of outputs (see \citealt{Bianchi2017} for a comprehensive overview). The most popular alternative to fully connected architectures are {\bf convolutional neural networks} (CNN), in which a much-limited set of connections exists between different layers: These connections perform a \emph{convolution} that gives the name of the architecture. Convolutional Neural Networks (CNNs) are predominantly used in image processing and applications that benefit from their advantageous balance between connectedness and complexity. However, CNNs require inputs to be structured in regular grids, such as those found in images. To generalize CNNs for data that is not available on a structured grid, Graph Neural Networks (GNNs) and more specifically, Graph Convolutional Networks (GCNs), have been developed \citep{Scarselli2009,Kipf2016}.

It is particularly instructive to unfold the recursive structure in \eqref{ANN} for the model in Figure \ref{ANN_example}. Both the recursive form and the scheme should be consulted in what follows. Starting from the last layer, we see that its neuron receives the output of three neurons from the previous layer, weighed by the {\bf connection weights} $w^{(l)}_{i,j}$, where $l$ denotes the layer hosting the neurons and the subscripts map the connection: for example, $w^{(3)}_{2,1}$ is the weight of the connection from neuron 2 (in layer 3) to neuron 1 (in layer 4). Following \eqref{ANN}, this neuron responds to these inputs as:

\begin{equation}
\label{eq17}
y=y^{(4)}=a^{(4)}\bigl(\sum^3_{j=1} w^{(3)}_{j,1} y_j^{(3)}+b^{(4)}_1\Bigr)=a^{(4)}\bigl (\bm{W}^{(3)}\bm{y}^{(3)} +b^{(4)}_1\bigr)\,.
\end{equation} 

The weight matrix for this layer is, in fact, a row vector $\bm{W}^{(3)}\in\mathbb{R}^{1\times 3}$ and the bias term is just a scalar $b^{(4)}_1\in\mathbb{R}$. The activation function could be classified into bounded and unbounded functions. The firsts are usually preferred in the last layers (near the output), while the first is generally more suited for the first layers (near the input). The tutorial in Section \ref{ch_3_sec_2} uses the hyperbolic tangent in the last layers and the ReLU (rectified linear unit) activation function in the others. These are defined as 

\begin{equation}
\label{eq18}
a(x)=\tanh(x) = \frac{e^x - e^{-x}}{e^x + e^{-x}} = \frac{1 - e^{-2x}}{1 + e^{-2x}} \quad ; \quad a(x)=\mbox{max}(0,x)
\end{equation}  

Note that a different activation function could be introduced for each neuron in a layer. This is the essence of Kolmogorov-Arnold networks (KANs, \citealt{Wang2024a}), which promise to significantly improve model efficiency. However, this approach increases the complexity of the structure and makes operations less amenable to parallelization via Graphical Processing Units (GPUs). KANs are currently one of the most exciting research avenues. Still, for this introduction, we stick to the traditional approach of using the same activation function for all neurons in a given layer.

Moving to the third layer, equation \eqref{ANN} gives 

\begin{equation}
\label{eq19}
\bm{y}^{(3)}= a^{(3)}\begin{bmatrix}
\Bigl(\sum^2_{j=1} w^{(2)}_{j,1} y_j^{(2)}+b^{(3)}_1\Bigr) \\
\Bigl(\sum^2_{j=1} w^{(2)}_{j,2} y_j^{(2)}+b^{(3)}_2\Bigr) \\ \Bigl(\sum^2_{j=1} w^{(2)}_{j,3} y_j^{(2)}+b^{(3)}_3\Bigr) \end{bmatrix}=a^{(3)}\Bigl (\bm{W}^{(2)}\bm{y}^{(2)} +\bm{b}^{(3)}\Bigr)\,,
\end{equation} with $\bm{W}^{(2)}\in\mathbb{R}^{3\times 2}$ and $\bm{b}^{(3)}\in\mathbb{R}^3$. Moving to the second layer, equation \eqref{ANN} sets 

\begin{equation}
\label{eq20}
\bm{y}^{(2)}= a^{(2)}\begin{bmatrix}
\bigl(w^{(1)}_{1,1} x+b^{(2)}_1\bigr) \\
\bigl(w^{(1)}_{1,2} x+b^{(2)}_2\bigr) \end{bmatrix}=a^{(2)}\bigl (\bm{W}^{(1)}x +\bm{b}^{(2)}\bigr)\,.
\end{equation} with $\bm{W}^{(1)}\in\mathbb{R}^{2\times 1}$ and $\bm{b}^{(2)}\in\mathbb{R}^2$.

% Comment for Jan: if this is the case
% lets just put it into the previous eq
%\begin{equation}
%\label{eq21}
%y_1^{(1)}=x\,%a^{(1)} (w_1^{1} x + b^{1}_1)\,
%\end{equation}

The reader is now encouraged to trace the full path from the input $x$ to the output $y$, inserting \eqref{eq20} into \eqref{eq19} and all the way up to \eqref{eq17}. It is evident that even a simple network with seven neurons embeds a cumbersome composite function:

\begin{equation}
\label{eq22}
y = a^{(4)}\bigl (\bm{W}^{(3)}  a^{(3)}\Bigl (\bm{W}^{(2)} a^{(2)}\bigl (\bm{W}^{(1)}x +\bm{b}^{(2)}\bigr) +\bm{b}^{(3)}\Bigr)  +b^{(4)}_1\bigr)\,.
\end{equation}

% \begin{equation}
% \label{eq22}
% y=a^{4}\Bigl (\bm{W}^{(3)}a^{(3)}\Bigl (\bm{W}^{(2)}a^{(3)}\Bigl (\bm{W}^{(1)}a^{(2)}\Bigl (\bm{W}^{(1)}a^{(1)} (w_1^{(1)} x + b^{1}_1) +\bm{b}^{(2)}\Bigr) +\bm{b}^{(3)}\Bigr) +b^{(4)}_1\Bigr)
% \end{equation}

In this simple architecture, the number of parameters (weight and biases) to be identified during the training amounts to $17$. Common architectures in deep learning have thousands of neurons and millions of parameters. For example, the famous AlexNet \citep{Krizhevsky2017} that revolutionized image classification and computer vision in 2012 is an ANN with 8 layers (5 convolutional and 3 feedforward) consisting of $65000$ neurons and $60$ million parameters. The training of this network took between five and six days using two GTX280 3GB GPUs and a training set of 15 million labeled images. This network significantly outperformed any classification strategy and set new standards in image classification. Yet, AlexNet is a toy compared to network architectures in modern Large Language Models (LLMs) such as GPT-4 or Jurassic-1 Jumbo, which have billions or even trillions of parameters.

The complexity of the nested architecture makes the ANNs training extremely challenging because of its many symmetries, which results in a vast amount of local minima \citep{Simsek2021}. Nevertheless, the most classic approach is gradient-based numerical optimization, with the gradient computed via back-propagation. The backpropagation algorithm was first proposed by \cite{Werbos}, reinvented several times and popularized by \cite{David}. A detailed derivation of the backpropagation algorithm is available in many sources \citep{Bishop1995,prince2023understanding,mendezHandsOnML} and is omitted here. On the other hand, gradient-based methods employed in modern deep learning libraries are so specific to their purpose that it is worth providing a short note about them.
Nearly all optimizers implemented for training ANNs are {\bf first order methods}. The reason is simple: most machine learning applications involve large datasets and large parameter space, hence the computation of the Hessian are usually too costly (if even possible) and impractical. The reader is referred to \cite{Yao2020,Anil2020} for a review of recent trends towards quasi-newton methods in deep learning.

The tutorials in the next chapter make use of the {\bf ADAptive Momentum estimation} (ADAM) optimizer to train an ANN and to drive a data assimilation algorithm. This combines ideas from momentum-accelerated gradient descent and gradient re-scaling. To understand its formulation, let us recall that the gradient descent algorithm for identifying the parameters $\bm{w}$ that minimize a cost function $\mathcal{J}(\bm{w})$ can be written as 

\begin{equation}
\label{GD}
\bm{w}^{(i+1)}=\bm{w}^{(i)}- \eta \frac{d\mathcal{J}}{d\bm{w}} (\bm{\Gamma}^{(i)},\bm{w}^{(i)}) \quad \eta  \in \mathbb{R}
\end{equation} where $\eta$ is the user defined \emph{learning rate} and the index $i$ denotes the current iteration. In the {\bf mini-Batch Gradient Descend} (BGD), the gradient is computed using randomly chosen portions (batches) of the data (denoted as $\bm{\Gamma}^{(i)}$ in \eqref{GD}) and the number of iterations is defined in terms of \emph{epochs}. An epoch is a full pass over the entire dataset: if we have 1000 data points and work with batches of 100 samples, then 1 epoch consists of 10 iterations of \eqref{GD}.

A way to increase convergence is to add momentum. The simplest implementation, due to \cite{Polyak1964}, reads:

\begin{equation}
\label{Mom}
\begin{split}
\bm{w}^{(i+1)}&= \bm{w}^{(i)}+ \bm{m}^{(i)} \\ \bm{m}^{(i)}&=\beta \bm{m}^{(i-1)}-\eta \frac{d\mathcal{J}^{(i)}}{d\bm{w}} \quad \eta, m, \beta \in \mathbb{R}\,,
\end{split}
\end{equation} having shortened the notation as $d\mathcal{J}^{(i)}/d\bm{w}=d\mathcal{J}/{d\bm{w}} (\bm{\Gamma}^{(i)},\bm{w}^{(i)}) $.

The parameter $\beta$ acts as a momentum/friction control which varies between 0 (high `friction') and $1$ (no `'friction') since the gradient now acts as an `acceleration' and not as a `velocity', the algorithm tends to go faster and uses inertia to escape from plateaus.

The main limitation of these approaches is that the learning rate is constant. A way to allow for faster steps on parameters that vary more slowly is to use the gradient re-scaling to give more "push" to parameters for which the gradient is small. The most popular approach is the RMSprop proposed by G. Hinton in his course on neural networks\footnote{Curiosity: this algorithm remains unpublished and is cited by the community as ``slide 29 in lecture 6''! This great lecture is available at \url{https://www.youtube.com/watch?v=defQQqkXEfE}.}. This algorithm introduces a scaling of the gradient such that 

\begin{equation}
\label{eq16}
\begin{split}
\bm{w}^{(i+1)}&= \bm{w}^{(i)}-\frac{\eta}{\sqrt{\bm{s}^{(i)}+\varepsilon}}\frac{d\mathcal{J}^{(i)}}{d\bm{w}}  \, \\  \bm{s}^{(i)} &= \beta \bm{s}^{(i-1)}+(1-\beta)\Biggl(\frac{d\mathcal{J}^{(i)}}{d\bm{w}}\Biggl)^2 \quad \eta, \beta, \epsilon \in \mathbb{R}\,.
\end{split}
\end{equation}

Here $\beta$ acts as a decay rate, $\bm{s}$ is a scaling vector and $\varepsilon$ is a small term introduced to avoid division by zero. The idea of this re-scaling is to decrease the learning rate faster for steepest directions while the parameter $\beta$ gives importance only to recent updates when computing $\bm{s}$. The ADAM optimizer combines \eqref{eq16} and \eqref{Mom} and reads:

\begin{align}
\bm{w}^{(i+1)} &= \bm{w}^{(i)}
   - \frac{\eta \hat{\bm{m}}}{\sqrt{\hat{\bm{s}}^{(i)}+\varepsilon}} \notag\\
\bm{s}^{(i+1)} &= \beta \bm{s}^{(i)}
   + (1-\beta)\!\left(\frac{d\mathcal{J}^{(i)}}{d\bm{w}}\right)^{\!2} \notag\\
\bm{m}^{(i+1)} &= \beta_1 \bm{m}^{(i)}
   + (1-\beta_1)\frac{d\mathcal{J}^{(i)}}{d\bm{w}} \notag\\
\hat{\bm{m}} &= \frac{\bm{m}}{1-(\beta_1)^{i}}, \qquad
\hat{\bm{s}} = \frac{\bm{s}}{1-(\beta_2)^{i}}, \qquad
\eta, \beta, \beta_1, \beta_2, \varepsilon \in \mathbb{R}.
\label{ADAM}
\end{align}

The idea is to have two moving averages: one for the squared gradient (like in RMSprop) and one for the momentum update. Although the number of tuning parameters has increased to four, it is rarely necessary to go beyond the classic default values. The implementation of \eqref{ADAM} in Python is particularly straightforward. A Python function implementing the optimizer using $numpy$ is provided for the exercise in the next chapter.

\subsection{Methods for non-parametric regression}\label{ch_2_sec_2_p_2}

Non-parametric methods do not rely on a pre-defined functional form. The two main categories are kernel-based methods and symbolic regression.

\paragraph{Kernel based methods} These methods are based on some measure of similarity between new inputs and those available in the training data. Therefore, contrary to parametric methods, these require storing the data in memory to make predictions.

Considering the case of a function $\tilde{f}:\mathbb{R}\rightarrow\mathbb{R}$, and a training dataset $\boldsymbol{\Gamma_*} = (\bm{x}_*,\bm{y}_*)$, we denote predictions at unseen points $\bm{x}^{**}$ as $\tilde{f}(x^{**}\mid \Gamma_*)$. The simplest example of non-parametric regression is linear interpolation. This proceeds in two phases: (1) identify the two nearest data points surrounding the query location, and (2) interpolate between them. For $\tilde{f}:\mathbb{R}\rightarrow\mathbb{R}$, if $x_1$ and $x_2$ are the closest distinct training inputs (assumed without loss of generality to satisfy $x_1 < x_2$), the prediction is
\begin{equation}
y = y_1 + \frac{y_2 - y_1}{x_2 - x_1}\,(x - x_1).
\end{equation}
It is often useful to write this in the equivalent barycentric form,
\begin{equation}
y = \left(\frac{x_2 - x}{x_2 - x_1}\right)y_1
    + \left(\frac{x - x_1}{x_2 - x_1}\right)y_2,
\end{equation}
which makes it clear that the prediction is a linear combination of the two nearest neighbors.

More generally, this idea extends to methods that use a larger number of neighbors or nonlinear weighting functions. A natural extension is the $k$-Nearest Neighbors (kNN) algorithm, which forms predictions based on the $k$ closest data points \citep{murphy2012machine}. Such approaches are sometimes referred to as \emph{instance-based} or \emph{lazy learning} methods, because no explicit training phase takes place: the model effectively ``memorizes'' the data and performs computation only at evaluation time.

\begin{comment}
$$\kappa(\bm{x}_i,\bm{x}_j)=\exp\Bigl (-\gamma ||\bm{x}_i-\bm{x}_j||^2 \Bigr)$$
\end{comment}

A variant of these methods that still requires a sort of training phase is the class of \emph{kernel methods}. A general template, considering for simplicity $\tilde{f}: \mathbb{R}\rightarrow \mathbb{R}$, reads

\begin{equation}
\label{non_parametric}
\bm{y}(\bm{x}_{**}| \bm{\Gamma}_*)=\sum^{n_{**}-1}_{j=0} \kappa(\bm{x}_{**,j},\bm{x}_*) \alpha_j = \bm{K}(\bm{x}_{**},\bm{x}_*) \bm{\alpha}\,,
\end{equation} where $\bm{x}_{**}\in\mathbb{R}^{n_{**}}$ is the set of new inputs, $\bm{\Gamma_*}=(\bm{x}_*,\bm{y}_*)$ is the training data and $\bm{K}(\bm{x}_{**},\bm{x}_*)\in\mathbb{R}^{n_{**}\times n_{*}}$ is the kernel matrix that measures the proximity between $\bm{x}_{**}$ and $\bm{x}_{*}$. The vector of parameters $\bm{\alpha}$ plays a similar role to the weight vectors. Although $\bm{\alpha}$ also needs to be identified in a sort of training phase, the training data remain necessary to evaluate the kernel matrix and thus to make predictions. The most popular kernel methods are (1) Kernel Ridge Regression, (2) Support Vector Machines and (3) Gaussian Process Regression. These arise from vastly different frameworks, but all fit in the template in \eqref{non_parametric} and solely differ in how $\bm{\alpha}$ is computed (see \citet{Kramer2013} for an overview). 

The Kernel Ridge Regression (KRR) can be derived form the \emph{kernelization} of the Ridge regression in \eqref{ridge} and provides a close form solution for $\bm{\alpha}$. To derive it, introduce \eqref{ridge} into \eqref{linear_parametric} to obtain: 

\begin{equation}
\label{smoo}
\bm{y}(\bm{x}_{**})=\bm{\Phi}(\bm{x}_{**})\bigl(\bm{\Phi}^T(\bm{x}_{*})\bm{\Phi}(\bm{x}_{*})+\alpha\bm{I}\bigr)^{-1}\bm{\Phi}^T(\bm{x}_{*})\bm{y}_*\,.
\end{equation}

Now use the matrix inversion lemma\footnote{This is also known as Woodbury matrix identity.} for the basis matrix $\bm{\Phi}(\bm{x}_{*})$:
\begin{equation}
\label{Wood}
\bigl ( \bm{\Phi}^T(\bm{x}_*)\bm{\Phi}(\bm{x}_*)+\alpha \bm{I}_{n_b}\bigr)^{-1}  \bm{\Phi}^T(\bm{x}_*)=\bm{\Phi}^T(\bm{x}_*)\bigl ( \bm{\Phi}(\bm{x}_*)\bm{\Phi}(\bm{x}_*)^T+\alpha \bm{I}_{n_*}\bigr)^{-1}\,.
\end{equation}

Introducing this identity into \eqref{smoo} gives 

\begin{equation}
\label{smoo2}
\bm{y}(\bm{x}_{**})=\bm{\Phi}(\bm{x}_{**})\bm{\Phi}^T(\bm{x}_*)\bigl ( \bm{\Phi}(\bm{x}_*)\bm{\Phi}(\bm{x}_*)^T+\alpha \bm{I}_{n_*}\bigr)^{-1}\bm{y}_*\,.
\end{equation}

%These two derivations can be linked to the notion of \emph{dual} problems in convex optimization, in that these are solution of two equivalent (dual) optimization problems.

The first important difference between \eqref{smoo} and \eqref{smoo2} is that \eqref{smoo2} requires inversion of matrices of size $n_{*}\times n_{*}$ rather than $n_b\times n_b$. The second is that all the matrices appearing have the form "$\bm{\Phi}\bm{\Phi}^T$". These can be seen as inner products between the rows of $\bm{\Phi}$. For certain choices of the bases \citep{bishop2006pattern}, these can be replaced by a kernel function that avoids the need for taking the inner products.
We can thus introduce the kernel function evaluation between two vectors $\bm{x}_1\in\mathbb{R}^{n_1}$ and $\bm{x}_2\in\mathbb{R}^{n_2}$ and the associated kernel matrix:

\begin{equation}
\bm{K}(\bm{x}_1,\bm{x}_2) = \bm{\Phi}(\bm{x}_1)\bm{\Phi}^T(\bm{x}_2) \in \mathbb{R}^{n_1\times n_2} \,.
\end{equation}

The term \emph{kernelization} refers to the process of replacing inner products with a kernel matrix computed with the appropriate kernel function.
Introducing the kernel formalism in \eqref{smoo2} closes the gap towards the template in \eqref{non_parametric}:

\begin{equation}
\label{smoo3}
\bm{y}(\bm{x}_{**})=\bm{K}(\bm{x}_{**},\bm{x}_*)\bigl ( \bm{K}(\bm{x}_{*},\bm{x}_*)+\alpha \bm{I}_{n_*}\bigr)^{-1}\bm{y}_*\,= \bm{K}(\bm{x}_{**},\bm{x}_*) \bm{\alpha}.
\end{equation} with

\begin{equation}
\label{alphav}
\bm{\alpha}=\bigl ( \bm{K}(\bm{x}_{*},\bm{x}_*)+\alpha \bm{I}_{n_{*}}\bigr)^{-1}\bm{y}_{*}\,.
\end{equation}

The reader is referred to \cite{murphy2012machine,Hastie2009,Welling,bishop2006pattern} for more details on the KRR. The key difference between KRR and Support Vector Regression (SVR) is that the latter is derived from a different cost function, which includes the idea of $\varepsilon$ sensitiveness in \eqref{E_epsilon} (see \cite{Smola2004sac} for a detailed tutorial). Identifying the parameters $\bm{\alpha}$ therefore requires numerical optimization and is generally more expensive. However, the benefit is that the result is usually a much sparser $\bm{\alpha}$, translating into much faster predictions. Moreover, the $\varepsilon$ sensitive functions make the SVR significantly more robust to outliers than KRR.

Finally, the same algebraic form used in Kernel Ridge Regression also appears in Gaussian Process Regression (GPR), but GPR is derived from a probabilistic viewpoint. A \emph{Gaussian process} is a distribution over functions characterized by a mean function and a covariance (kernel) function, such that any finite collection of function values has a joint multivariate Gaussian distribution \citep{Mendez2022}. In GPR, we assume that the unknown function is drawn from such a Gaussian process, with the covariance function playing the role of the kernel \citep{CarlEdward}. The prediction formula \eqref{non_parametric} is then obtained by conditioning this Gaussian process on the observed training data. For functions $f:\mathbb{R}^{n_x}\rightarrow\mathbb{R}$, a set of $n$ predictions $\bm{y}$ evaluated at inputs $\bm{x}$ is thus jointly distributed with the training data $(\bm{x}^*,\bm{y}^*)$ according to:

\begin{equation}
\label{GAUSS}
\begin{pmatrix}
    \mathbf{y}_* \\
    \mathbf{y}
\end{pmatrix}
\sim \mathcal{N} \left(
    \bm{0}, 
    \begin{pmatrix}
        \mathbf{K}_{**} & \mathbf{K}_*^T \\
        \mathbf{K}_* & \mathbf{K}
    \end{pmatrix}
\right)
\end{equation} where $\bm{K}_{**}=\kappa(\bm{x}_*,\bm{x}_*)\in\mathbb{R}^{n_*\times n_*}$ is the covariance matrix of the training data, $\bm{K}=\kappa(\bm{x},\bm{x})\in\mathbb{R}^{n\times n}$ is the covariance matrix of the new sample points and $\bm{K}_{*}=\kappa(\bm{x},\bm{x}_*)\in\mathbb{R}^{n\times n_*}$ is the cross-covariance matrix between new inputs and training inputs. The transpose $\mathbf{K}_*^{\mathsf{T}}$ therefore represents the cross-covariance in the opposite direction. Note that the assumption of zero means $\bm{0}\in\mathbb{R}^{(n+ n_*)}$ in \eqref{GAUSS} is merely a matter of convenience since adding a more complex assumption does not generally pay out in terms of added model capacity. The predictions $\bm{y}$ also constitute a multivariate Gaussian, whose mean and covariance can thus be computed from \eqref{GAUSS} via conditioning. 
The derivation of the formulae is left as an exercise\footnote{%
\setlength{\parskip}{0.4ex}%
\setlength{\parindent}{0pt}%
Recall that, given two random variables $\bm{x}_a$ and $\bm{x}_b$, jointly distributed according to a multivariate Gaussian of the form:
\begin{center}
\begin{minipage}{0.9\linewidth}
\[
\mathbf{x} =
\begin{bmatrix}
    \mathbf{x}_a \\
    \mathbf{x}_b
\end{bmatrix}
\sim \mathcal{N}\!\left(
    \begin{bmatrix}
        \mu_a \\
        \mu_b
    \end{bmatrix},
    \begin{bmatrix}
        \Sigma_{aa} & \Sigma_{ab} \\
        \Sigma_{ab}^\top & \Sigma_{bb}
    \end{bmatrix}
\right)
\]
\end{minipage}
\end{center}
we have that the conditioning is also Gaussian:
\[
p(\bm{x}_a|\bm{x}_b) \sim \mathcal{N}(\bm{\mu}_{a|b}, \bm{\Sigma}_{a|b})
\]
with
\[
\bm{\mu}_{a|b} = \bm{\mu}_a + \bm{\Sigma}_{a,b} \bm{\Sigma}^{-1}_{b,b} (\bm{x}_b - \bm{\mu}_b),
\qquad
\bm{\Sigma}_{a|b} = \bm{\Sigma}_{a,a} - \bm{\Sigma}_{a,b} \bm{\Sigma}^{-1}_{b,b} \bm{\Sigma}_{b,a}.
\]
}.

The Gaussian Process regression is the multivariate Gaussian obtained by conditioning $p(\bm{y}_*|\bm{y})$ in \eqref{GAUSS}.
Interestingly, it is possible to arrive at the same result using a Bayesian KRR formulation and the matrix inversion lemma (the complete derivation is given in \citet{mendezHandsOnML}).

%and the mean prediction is identical to the KRR \citep{gundersen2020kernelgp}. However, the evaluation of uncertainties is slightly different and generally more conservative than in KRR \citep{mendezHandsOnML}.
\vspace{-5mm}
\paragraph{Symbolic Regression.}Symbolic regression (see \citet{la2021contemporary,udrescu2020ai,Schmidt2009science}) consists in identifying a symbolic expression (mathematical formula) that best fits the given data. Contrary to the other approaches, a prescribed parametric shape is not provided. Instead, a set of possible function candidates is defined and the optimization algorithm is free to modify it within certain limits.

The optimization is generally carried out using evolutionary algorithms, the most popular one being the \emph{genetic programming} \citep{Koza1994,Banzhaf1998book}. Genetic Programming (GP), developed by Koza \citep{Koza1994} as a new paradigm for automatic programming and machine learning \citep{Banzhaf1998book,Vanneschi2012} is able to optimize both the structure and the parameters of a model.

\begin{figure}[htpb]
\center
 \includegraphics[width=0.5\textwidth]{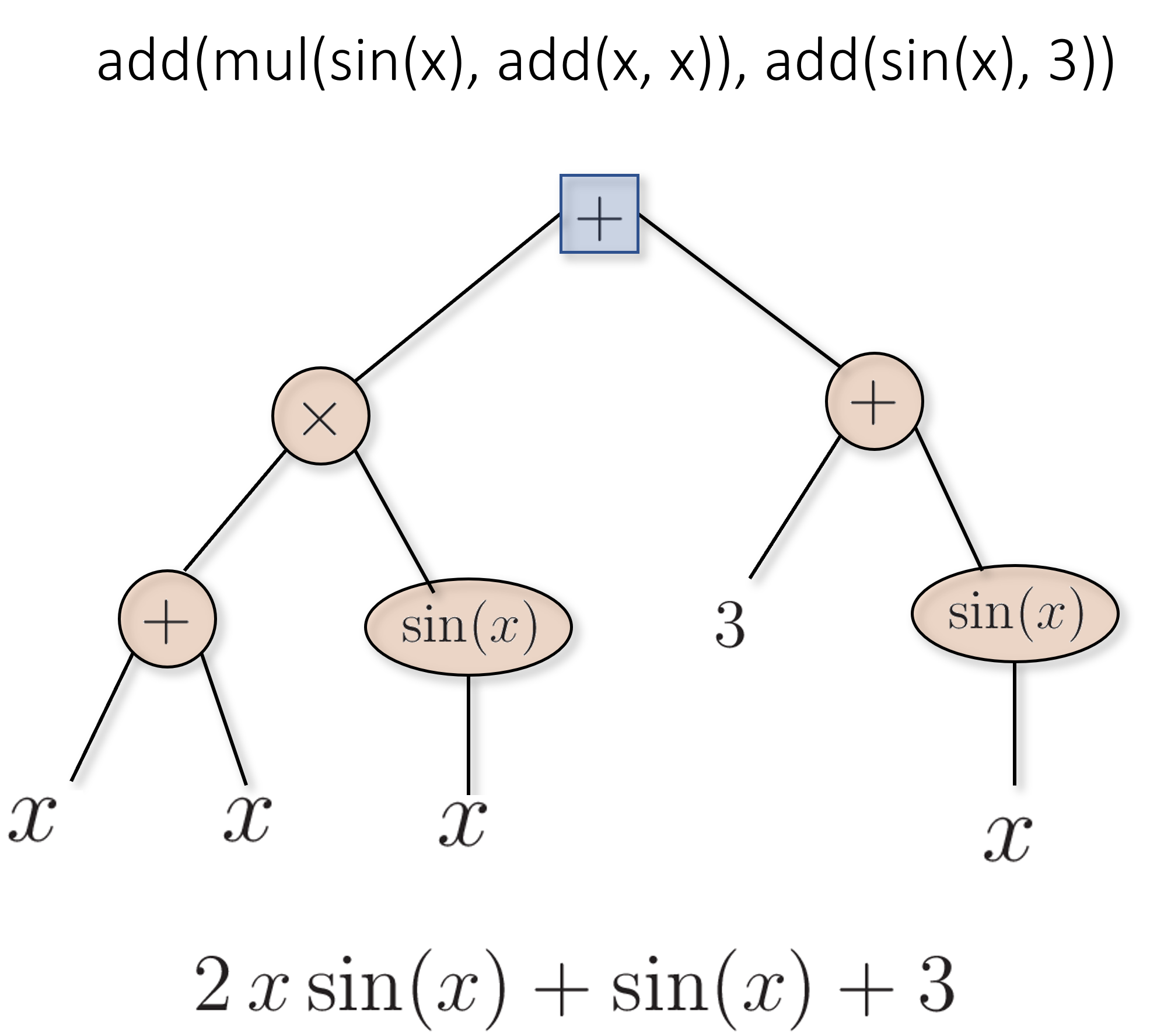}
    \centering
    \caption{Syntax tree representation of the function $2 x\sin(x)+\sin(x)+3$. This tree has a root '+' and a depth of two. The nodes are denoted with orange circles, while the last entries are leafs.}
    \label{GP_Tree}
\end{figure}

The parametric function takes the form of recursive trees of predefined functions connected through mathematical operations. These trees are encoded into a string, which includes arithmetic operations, mathematical functions, Boolean operations, conditional operations, or iterative operations. An example of a syntax tree representation of a function is shown in Figure \ref{GP_Tree}. A tree (or \emph{program} in GP terminology) is composed of a root that branches out into nodes (containing functions or operations) throughout various levels. The number of levels defines the \emph{depth} of the tree, and the last nodes are called \emph{terminals} or \emph{leaves}. These contain the input variables or constants. Any combination of branches below the root is called \emph{sub-tree} and can generate a tree if the node becomes a root.

The user specifies the \emph{primitive set}, i.e. pool of allowed functions, maximum depth of the tree, etc. The GP then operates on a population of possible candidate solutions (individuals) and evolves it over various steps (generations) using genetic operations in the search for the optimal tree.
Classic genetic operations include elitism, replication, cross-over and mutations, as in Genetic Algorithm Optimization \citep{haupt2004practical}. 
A popular open-source Python library for symbolic regression with genetic programming is DEAP \citep{DEAP_JMLR2012}.

\section{Data driven... Scientific computing} \label{ch_2_sec_3}

Several important parallelisms can be drawn between the mathematical framework and the set of operations underlying the process of training a parametric model and the process of solving a Partial Differential Equation (PDE) using numerical methods.

Let us provide a very high-level overview of the general procedure for numerically solving a PDE using finite element methods (FEM), arguably the most general-purpose approach for the task \citep{whiteley2014finite,solin2005partial}. Consider a general PDE operator $\mathcal{D}$ that involves various partial derivatives of a scalar function $f:\bm{x}:=(x,y):\rightarrow z \in\mathbb{R}$ in a domain $\Omega\subset \mathbb{R}^2$ with boundaries $\partial \Omega$ in which a set of boundary conditions $\mathcal{B}$ is imposed:
\begin{subequations}
    \label{PDE}
\begin{equation}
\mathcal{D} (f, \partial_{x} f, \partial_{y} f, \partial_{xy} f, \partial_{xx} f \dots)= 0\,
\end{equation}
\begin{equation}
\mathcal{B}(f(\bm{x}_{B}),\partial_n f(\bm{x}_{N}))=0
\end{equation}
\end{subequations}

Numerical methods seek a numerical approximation of the function $f$. The general approach in FEM is to look for an approximation to such a function written in the form of a linear combination of $n_m$ basis functions, i.e. :

\begin{equation}
\label{basis_FEM}
f(\bm{x})\approx \tilde{f}(\bm{x})=\sum^{n_m-1}_{j=0} \phi_j (\bm{x}) f_j =\tilde{f}(\bm{x};\bm{f})  \,.
\end{equation}

\begin{figure}\center
 \includegraphics[width=0.85\textwidth]{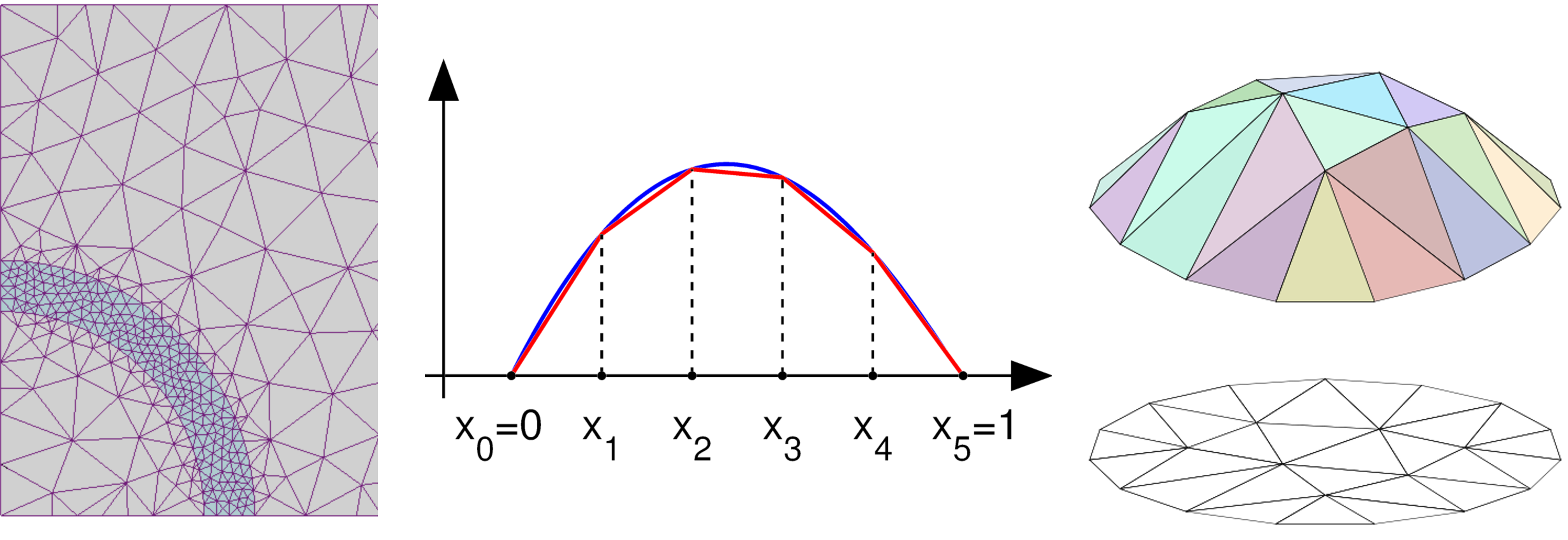}
    \centering
    \caption{The numerical discretization in a FEM approach to the numerical solution of a PDE. A computational mesh (left) is used to collocate the basis functions in \eqref{basis_FEM}. These functions are used for interpolation (middle) rather than regression, i.e., one expects $f(\bm{x}_j)=\bm{f}_j$. A linear combination of these bases can be used to approximate a surface/function (right) that solves a PDE.}
    \label{Fig_FEM}
\end{figure}

These bases are \emph{local}, in the sense that they are different from zero only in a small portion of the domain, similarly to the RBFs. However, unlike the RBFs, their definition requires the formulation of a \emph{mesh} that discretizes the domain, and their role is to provide an interpolation and not a regression. Every basis element equals 1 at the center of the considered mesh cell and is zero in all the other centers. This ensures that one has $f(\bm{x}_j)=\bm{f}_j$ for all j, which is generally not the case and not a desirable condition in a regression. Figure \ref{Fig_FEM} illustrates the key ingredients.

In its most general formulation, the FEM does not solve \eqref{PDE} directly, but instead projects the problem onto a set of \emph{test functions} $v_k(\bm{x})$, yielding the so-called \emph{weak form} of the PDE. This approach provides several important advantages (see \citet{whiteley2014finite,solin2005partial}), including a systematic way to incorporate boundary conditions. The result is a system of $n_m$ equations of the form
\begin{equation}
\int_{\Omega} v_k(\bm{x})\, \mathcal{D}\!\left(\tilde{f}, \partial_{x}\tilde{f}, \partial_{y}\tilde{f}, \partial_{xy}\tilde{f}, \partial_{xx}\tilde{f}, \dots \right)\, \mathrm{d}\bm{x}
= R_{\mathcal{D}k}(\bm{f}), \qquad k = 0,\dots,n_m-1,
\end{equation}
where the terms on the right-hand side, which we aim to drive to zero (or as close to zero as possible), are called \emph{residuals}. Collecting them forms a residual vector $\bm{r}_{\mathcal{D}}$, and solving the PDE numerically amounts to finding the coefficient vector $\bm{f}$ that minimizes this residual. In what follows, we refer to the associated objective as a \textbf{differential cost function} $\mathcal{R}_{\mathcal{D}}(\bm{f})$, since it involves derivatives of the function being approximated; more generally, we describe it as a \textbf{physics-driven} cost function, in contrast to the \textbf{data-driven} cost functions introduced earlier.

At this high level, the problem resembles the training of a parametric model in machine learning, as it involves (1) choosing a parametric representation of the solution, (2) defining a performance measure to evaluate a candidate solution, and (3) applying an optimization method to iteratively improve it.

One of the most promising avenues of the field is to take advantage of this parallelism to promote a fusion between data-driven science and computational engineering. One could, for example, use parametric functions from machine learning to solve PDEs or add the minimization of differential cost functions as additional requirements for the training of machine learning models. The problem of combining two objective functions (e.g., the general data-driven $\mathcal{J}$ with the general physics-driven $\mathcal{R}_{\mathcal{D}}(\bm{w})$) can be viewed as a problem in multidisciplinary or in constrained optimization. 

\begin{enumerate}
\item \textbf{Architecturally Constrained Parametric Model}. The most natural and yet robust approach is to design a parametric function $\tilde{f}(\bm{x};\bm{w})$ that structurally complies with the physics-based constraints. This can be done at the level of the input definition (feature selection, in the machine learning terminology) and at the level of model architecture. Valid inputs are often dimensionless numbers that respect the scaling laws of the problem \citep{Dominique2022,Calado2023}. In terms of model definition, one might combine machine learning models with carefully engineered models that embed physical principles. These could be placed downstream the prediction chain. For example, if one seeks to model the mean flow field in a channel to infer data-driven turbulence models, a simple solution to ensure that the prediction complies with the non-slip conditions at the wall could be to multiply the model prediction with a function that is zero on the walls. A more sophisticated example is provided in \cite{Fiore2022}, where the developed data-driven model for turbulent heat flux structurally complies with the Galilean invariance and the second law of thermodynamics.

%$$\bm{y}=f(\bm{x}, m(\check{\bm{x}};\bm{w}))$$

%$$\mathcal{R}_{D}(\bm{w})$$

This approach should always be considered, as it greatly enhances the robustness of a data-driven model at the cost of minor modifications of the training pipeline. On the other hand, from a modeling perspective, approaches of this kind often trade robustness with generalization since ad hoc parametrization is often only valid for specific conditions. Moreover, their design requires significant domain knowledge and expertise. We see examples of this approach in the tutorials of Chapter \ref{chapter3}

\item \textbf{Penalizations and Regularizations}. This is the simplest and most popular approach. The idea is to combine the data-driven and the physics-driven cost functions in a single cost function as $\mathcal{A}= \mathcal{J}+\alpha \mathcal{R}_D$, with $\alpha\in\mathbb{R}^+$ a user-defined parameter controlling the relative importance of the second term over the first one. This approach requires no modification to the training of methods such as genetic programming and only minor modifications to the training of methods such as RBFs or ANNs, provided that the gradient $d\mathcal{R}_D/d{\bm{w}}$ can be easily computed.

This idea is largely exploited in the popular Physics Informed Neural Networks (PINNs), which uses ANNs to solve ODEs and PDEs. These were first introduced by \cite{psichogios1992hybrid}, further developed by \cite{lagaris1998artificial} (who referred to the approach as ``hybrid neural-network-first principle modeling''), and then popularized and adapted to modern python libraries by \cite{raissi2019physics} (who coined the term ``PINNs''). Today, many powerful libraries implementing PINNs are open-source and continuously under development \citep{lu2021deepxde,Haghighat2021,Coscia2023,Peng2021IDRLnetAP}.

Although easy to set up and train, the penalized method requires the user definition of the penalty (or penalties, if more physics-driven conditions are requested). It is often difficult to define this parameter because it is challenging to ensure that the two terms ($\mathcal{J}$ and $\alpha \mathcal{R}_D$) are treated ``equally'' by the training optimizer. Even when an appropriate balance is reached, the optimal solution usually seeks a compromise that does not ensure the fulfillment of the condition to machine precision. This can be problematic for problems extremely sensitive to, e.g., boundary conditions. A remedy is the use of methods for re-scaling\footnote{A comprehensive and didactic talk concerning this problem is provided by Paris Perdikaris and is available, at the time of writing, at \url{http://www.ipam.ucla.edu/abstract/?tid=15853&pcode=MLPWS3}.} the gradient of the cost function during the training \citep{Wang2021}. 

In summary, penalties are usually of great help and always worth considering, given how simple it is to set them up. However, one should not solely rely on these to fully enforce the physics-driven information unless fairly sophisticated methods are used.

\item \textbf{Lagrange Multipliers and Hard Constraints.} 
If the previous approach can be viewed as adding \emph{soft} constraints, this framework enforces \emph{hard} constraints. The training problem is formulated as a constrained optimization, where the data-driven cost function $\mathcal{J}$ is minimized subject to the physics-driven constraint $\mathcal{R}_{\mathcal{D}} = 0$. The literature on constrained optimization is extensive (see \citet{NOCEDAL_2006a} and \citet{Martins2021} for overviews), and many algorithmic strategies are available.

The general idea is to introduce the augmented function $\mathcal{A} = \mathcal{J} + \bm{\lambda}^\mathsf{T} \mathcal{R}_{\mathcal{D}}$, where $\bm{\lambda}\in\mathbb{R}^{n_f}$ is the vector of Lagrange multipliers and $n_f$ is the number of constraints. Unlike the soft-constraint approach, this formulation requires solving for both $\bm{f}$ and $\bm{\lambda}$, making the problem numerically more involved.

For linear methods such as Radial Basis Functions (RBFs), this constrained formulation recovers well-known structures when addressing classical PDEs. When enforcing PDE constraints together with standard boundary conditions (e.g., Dirichlet or Neumann), the resulting system often reduces to a quadratic objective with linear constraints, leading to a large linear system \citep{Sperotto2022a}. More broadly, the use of RBF expansions to solve PDEs without a computational mesh dates back to \citet{Kansa1990,Kansa1990a}, and has since generated a substantial literature (see, e.g., \cite{Fornberg2015,cmes,Chen2002,Chen2003,Sarler}). RBF-based meshless methods extend classical pseudo-spectral approaches \citep{Fornberg1996}, where Fourier or Chebyshev expansions are typically used, and can be interpreted as a class of collocation schemes. \emph{Arguably}, one of the main reasons these methods have not achieved the same widespread adoption as FEM is that the resulting linear systems tend to be significantly less sparse, and therefore more memory-intensive.

For nonlinear methods, the constraining leads to less explored territory. To the authors' knowledge, no attempt has been made to combine a fully constrained formalism with Genetic Programming for solving PDEs, at least for fluid dynamic applications, and all known approaches in this direction rely on a penalization framework (see \cite{Tsoulos2006,Sobester2008,Pratama2023,Oh2023a}. Concerning constrained ANNs, this is arguably the most promising and recent avenue. The first approach was recently proposed by \cite{Basir_2022} (see also \cite{Basir2023} and \cite{Son2023}). Much development can be expected soon. 

\end{enumerate}

\section{Summary and Conclusions}\label{ch_2_sec_4}

This chapter provided a broad overview of regression methods in machine learning and of strategies for incorporating physics-based information into the learning process. We began by framing regression as the task of fitting not just a single curve, but a stochastic process—a distribution of possible functions—to observed data. The simplest viewpoint treated the function as the sum of a deterministic component and a zero-mean stochastic term. We then moved to a probabilistic interpretation, showing how different assumptions on the stochastic term lead to different cost functions, which we referred to as \emph{data-driven} cost functions. We concluded this part by introducing bootstrapping and cross-validation, fundamental tools for assessing generalization performance and understanding the impact of limited data.

We then contrasted the data-driven learning framework with the classical setting of \emph{scientific computing}, where the target function is the solution of a physics-based model, typically expressed as a PDE. Drawing parallels between training parametric models and solving PDEs numerically allowed us to outline methods that combine these two perspectives: seeking functions that both match the available data \emph{and} satisfy the governing physical laws. 

With this foundation in place, we are now prepared to move to the next chapter, which presents three tutorial exercises illustrating practical approaches to such hybridization.

%\bibliographystyle{apalike}
%\bibliography{chap2/chap2bib}

\bibliographystyle{apalike}

\bibliography{chapter_2}

% USER ENTRY OFF
%\clearpage{\pagestyle{empty}\cleardoublepage}
% USER ENTRY ON
%\input{appendix}
% comment this input if you do not have an appendix
% USER ENTRY OFF
\end{document}